\documentclass[letterpaper]{article} 
\usepackage{aaai25}   
\usepackage{times}  
\usepackage{helvet}  
\usepackage{courier}  
\usepackage[hyphens]{url}  
\usepackage{graphicx} 
\usepackage{natbib}  
\usepackage{caption} 
\usepackage{algorithm}
\usepackage{algorithmic}
\usepackage{amsmath}
\usepackage{amssymb} 
\usepackage{adjustbox}
\usepackage{threeparttable}
\usepackage{amsfonts}
\usepackage[bb=dsserif]{mathalpha}
\usepackage{multirow}
\usepackage{colortbl}
\usepackage{tcolorbox}
\definecolor{light-gray}{gray}{0.94}
\usepackage{cleveref} 
\usepackage{tikz}
\usepackage{booktabs}
\usepackage{multirow}
\def\loss{{\mathcal{L}}}

\DeclareMathOperator*{\argmin}{arg\,min}

\usepackage{newfloat}
\usepackage{listings}
\DeclareCaptionStyle{ruled}{labelfont=normalfont,labelsep=colon,strut=off} 
\floatstyle{ruled}
\newfloat{listing}{tb}{lst}{}
\floatname{listing}{Listing}
\title{Toward Adaptive Large Language Models Structured Pruning via \\ Hybrid-grained Weight Importance Assessment}
\author{
    Jun Liu\textsuperscript{\rm 1,2}, 
    Zhenglun Kong\textsuperscript{\rm 1},
    Pu Zhao\textsuperscript{\rm 1},
    Changdi Yang\textsuperscript{\rm 1},
    Xuan Shen\textsuperscript{\rm 1},
    Hao Tang\textsuperscript{\rm 3,2}\thanks{Corresponding Authors.},,
    Geng Yuan\textsuperscript{\rm 4}, 
    Wei Niu\textsuperscript{\rm 4},
    Wenbin Zhang\textsuperscript{\rm 5},
    Xue Lin\textsuperscript{\rm 1},
    Dong Huang\textsuperscript{\rm 2}\footnotemark[1],
    Yanzhi Wang\textsuperscript{\rm 1}\footnotemark[1]
}
\affiliations{
    \textsuperscript{\rm 1}Northeastern University \\
    \textsuperscript{\rm 2}Carnegie Mellon University \\
    \textsuperscript{\rm 3}Peking University \\
    \textsuperscript{\rm 4}University of Georgia \\
    \textsuperscript{\rm 5}Florida International University 
   
}

\begin{document}

\maketitle

\renewcommand{\thefootnote}{\fnsymbol{footnote}}
\footnotetext[1]{Corresponding Authors.}

\begin{abstract}
Structured pruning for large language models (LLMs) has garnered significant academic interest due to its ability to efficiently compress and accelerate LLMs by eliminating redundant weight groups at a coarse-grained granularity. Current structured pruning methods for LLMs typically depend on a singular granularity for assessing weight importance, resulting in notable performance degradation in downstream tasks. Intriguingly, our empirical investigations reveal that utilizing unstructured pruning, which achieves better performance retention by pruning weights at a finer granularity, \emph{i.e.}, individual weights, yields significantly varied sparse LLM structures when juxtaposed to structured pruning. This suggests that evaluating both holistic and individual assessment for weight importance is essential for LLM pruning. Building on this insight, we introduce the Hybrid-grained Weight Importance Assessment (HyWIA), a novel method that merges fine-grained and coarse-grained evaluations of weight importance for the pruning of LLMs. Leveraging an attention mechanism, HyWIA adaptively determines the optimal blend of granularity in weight importance assessments in an end-to-end pruning manner. Extensive experiments on LLaMA-V1/V2, Vicuna, Baichuan, and Bloom across various benchmarks demonstrate the effectiveness of HyWIA in pruning LLMs. For example, HyWIA surpasses the cutting-edge LLM-Pruner by an average margin of 2.82\% in accuracy across seven downstream tasks when pruning LLaMA-7B by 50\%. Code:https://github.com/azuryl/LLM-HWIA

\end{abstract}

%

\section{Introduction}

Large Language Models (LLMs) have demonstrated unparalleled efficacy in various application domains~\cite{li2023textbooks, touvron2023llama,  chowdhery2023palm}.
However, deploying LLMs at inference time incurs significant financial and energy costs, mainly due to their large model scale, which requires extensive computational resources and GPU memory~\cite{zhao2023survey,shen2024search}.
In response, there has been marked increase in interest in compressing LLMs, which upholds the promise of LLMs while substantially reducing their memory requirements and computational costs.
Prominent techniques include parameter quantization~\cite{xiao2023smoothquant, shao2023omniquant}, network pruning~\cite{ frantar2023sparsegpt, yuan2021work, yuan2022mobile,zhao-etal-2024-pruning}, token reduction~\cite{zhan2024exploring,zhan-etal-2024-rethinking-token} and low-rank decomposition~\cite{bach2005predictive},~\emph{etc}.
\begin{figure}[t]
  \includegraphics[width=1\columnwidth]{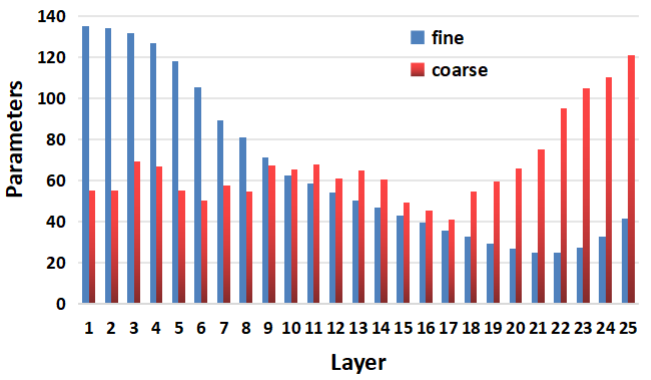}
  \caption{{Sparsity allocation across different layers of LLaMA-7B pruned by fine-grained ~\cite{xia2022structured} and coarse-grained~\cite{lee2020flexible} weight importance criteria (50\% global pruning rate). Fine-grained pruning tends to preserve more weight in the shallow layers, which is in stark contrast to coarse-grained pruning. The vertical axis represents the parameter quantity of each layer in terms of millions. The horizontal axis represents the layer number of LLaMA-7B. }}
  \label{fig:motive}
  \vspace{-0.7cm}
\end{figure}

This paper focuses on pruning LLMs by removing redundant parameters to produce a sparse, lightweight model. Pruning methods vary in granularity, ranging from fine- to coarse-grained approaches. Fine-grained pruning evaluates the importance of individual weights, as seen in SparseGPT~\cite{frantar2023sparsegpt}, which uses the Hessian matrix for layer-wise weight reconstruction, and Wanda~\cite{sun2023simple}, which combines weight magnitude with input activations to assess significance. While effective in reducing model size with minimal performance loss, fine-grained pruning creates irregular sparsity patterns, complicating deployment on conventional hardware.

In contrast, coarse-grained (structured) pruning eliminates entire columns, rows, or blocks of weights, leveraging metrics like gradient information~\cite{ma2023llm} for importance assessment. This approach simplifies deployment and achieves acceleration but often incurs a greater performance drop compared to unstructured pruning, even with fine-tuning~\cite{sun2023simple}.

Broadly speaking, current LLM structured pruning methods typically rely solely on a single granularity of weight importance assessment.  
Interestingly, we empirically observed that estimating weight importance across different granularities can produce markedly diverse sparse structures in LLMs. 
As illustrated in Figure~\ref{fig:motive}, fine-grained estimations prioritize the weights in the initial layers as most critical, thereby preserving a greater number of weights, while the coarse-grained counterparts exhibit the opposite tendency. 
Delving deeper, fine-grained estimation~\cite{han2015learning, frantar2023sparsegpt} focuses on sustaining and calculating the contribution of each weight to the network output.
In contrast, coarse-grained estimation~\cite{ma2023llm, zhang2023loraprune} predominantly considers the overall effect along weight groups, which may neglect the extreme values of individual weight that holds significance, ~\emph{i.e.}, weight outliers~\cite{xiao2023smoothquant}.
Therefore, how to simultaneously perceive and evaluate the importance of individual weights and holistic weight groups remains an unresolved challenge in the field.

To address these bottlenecks, we propose the Hybrid-grained Weight Importance Assessment (HyWIA), which adaptively integrates fine-grained and coarse-grained weight importance estimations. By leveraging the attention mechanism~\cite{vaswani2017attention}, HyWIA automatically generates hybrid-granularity importance scores.
This facilitates dynamic balancing and weighting of importance scores at various granularities, thus allowing for a more robust assessment of importance from both individual and collective weight group perspectives.
Comprehensive experiments on pruning a variety of LLMs including LLaMA~\cite{touvron2023llama}, Vicuna~\cite{vicuna2023}, Baichuan~\cite{yang2023baichuan}, and Bloom~\cite{workshop2022bloom}, demonstrate the superiority of HyWIA over many state-of-the-art methods. 
For example, HyWIA significantly enhances performance compared to LLM-pruner~\cite{ma2023llm} and LoRAPruner~\cite{zhang2023loraprune}, further improving accuracy by 2.82\% and 2.09\% respectively with LLaMA-7B at the 50\% pruning rate.
The main contribution of this paper can be summarized as:
\begin{itemize}
    \item We empirically observed that coarse-grained and fine-grained pruning generate markedly different sparsity distributions across LLM layers. This largely indicates that structured pruning methods overlook the importance assessment of individual weights, thereby explaining their performance deficit relative to unstructured pruning.
    \item We introduce HyWIA, a novel LLM pruning method that adaptively merges fine-grained and coarse-grained metrics to comprehensively assess the importance of weights. To the best of our knowledge, this is the first instance of proposing a hybrid-granularity assessment for weight importance in the community.
    \item Extensive experiments on pruning representative LLMs demonstrate the superiority of the proposed HyWIA over state-of-the-art methods.
\end{itemize}

\section{Background and Motivation}

Model pruning commonly comprises three steps~\cite{NIPS19896c9882bb,molchanov2016pruning,rtseg}. Recently, some researchers introduced grouping~\cite{ma2023llm, sun2023simple} as the first step, aiming to group structures within large models. The second step is the importance estimation step, during which redundant weight groups selected for pruning are identified. The third step, LoRA fine-tuning~\cite{kwon2022fast,ma2023llm,sun2023simple}, concludes the pruning process, aiming to quickly restore any performance that may have been affected by the removal of parameters.

\begin{figure*}[h]
  \centering
  \includegraphics[width=1\linewidth]{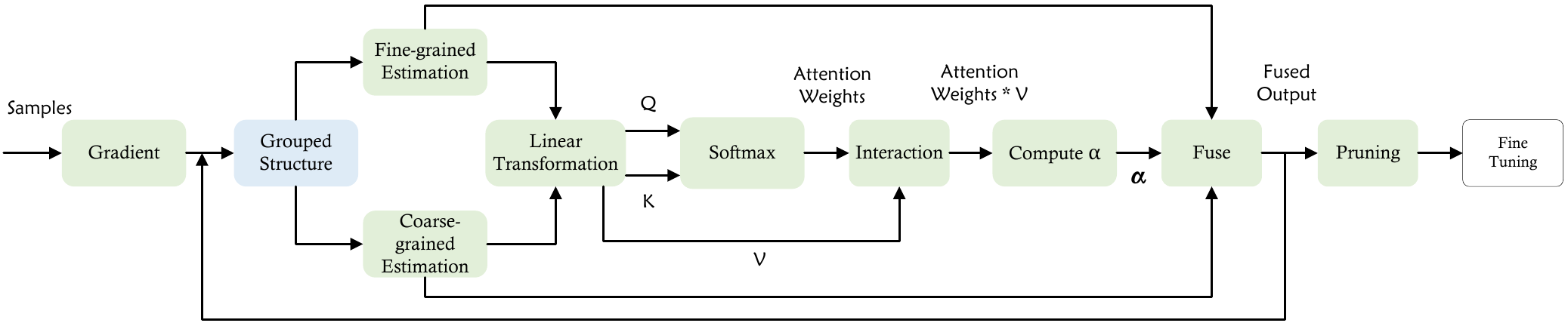} \hfill
  \caption{The framework of our proposed Hybrid-grained Weight Importance Assessment (HyWIA) consists of three stages: grouping (blue), adaptive estimation (green), and fine-tuning (white). In the grouping stage, we construct the dependency structure within the LLM. The adaptive estimation stage includes gradient calculation, fine-grained and coarse-grained importance estimation, adaptive fusion, element sorting, and pruning. Finally, the fine-tuning stage uses LoRA~\citep{hu2022lora} to recover the pruned model's performance and functionality.}
  \label{fig:suanfa}
    \vspace{-0.4cm}
\end{figure*}

\subsection{Problem Formulation}

The pruning problem is framed as an optimization problem, where the goal is to find an optimal mask \( \textbf{m} \) under a constraint.

Given a loss function \( \mathcal{L}(\textbf{m}) \) that depends on a mask \( \textbf{m} \) (where \( \text{m} \) determines which parameters are kept or pruned), the second-order Taylor series expansion around an initial mask \( \mathbb{1} \) (which typically represents keeping all parameters) is given by~\cite{NIPS19896c9882bb,kwon2022fast}:
\begin{equation}
\mathcal{L}(\text{m}) \approx \mathcal{L}(\mathbb{1}) + \nabla_{\text{m}} \mathcal{L}(\mathbb{1})^\top (\text{m} - \mathbb{1}) + \frac{1}{2} (\text{m} - \mathbb{1})^\top \mathbf{H} (\text{m} - \mathbb{1}),
\end{equation}
where:
\begin{itemize}
    \item \( \mathcal{L}(\mathbb{1}) \) is the loss at the initial mask.
    \item \( \nabla_{\text{m}} \mathcal{L}(\mathbb{1}) \) is the gradient of the loss with respect to the mask.
    \item \( \mathbf{H} \) is the Hessian matrix (second-order derivative) of the loss with respect to the mask.
\end{itemize}

Assuming the model is near a local minimum where the gradient is small to 0 we ignore the linear term, and as $\loss(\mathbb{1})$ is a constant, the optimization objective as follows:
\begin{equation}
    \argmin_{\text{m}} \loss(\text{m}) 
    \approx \argmin_{\text{m}} (\mathbb{1} - \text{m})^\intercal \text{H} (\mathbb{1} - \text{m}). \label{eq:argmin_hessian}
\end{equation}

Since directly computing the Hessian matrix \( \mathbf{H} \) is impractical, it is approximated by the empirical Fisher information matrix~\cite{kwon2022fast} \( \mathbf{F} \).

\noindent{\textbf{Motivation.}} In LLMs, the decoders situated in the initial layers possess distinctive parameters that wield a vital role in capturing intricate characteristics of the input tokens. Consequently, fine-grained estimation manifests as highly suitable for these layers. Conversely, the decoders occupying the final layers of LLMs prioritize the comprehension of semantics and context. Here, a specific coupled structure assumes a pivotal role in grasping abstract semantics and establishing long-distance dependency relationships. As a result, coarse-grained estimation emerges as particularly fitting for these layers. The current LLM method~\cite{frantar2023sparsegpt,ma2023llm,sun2023simple} only emphasizes general estimation methods such as fine-grained or coarse-grained, resulting in a limited holistic consideration that fails to integrate the strengths and advantages of both approaches. Consequently, challenges arise when estimating the importance of each layer.

We prune LLaMA-7B using fine-grained (Appendix C.1) and coarse-grained (Appendix C.2) estimation methods, each at a 50\% pruning rate. Figure~\ref{fig:motive} shows that fine-grained pruning retains more parameters in the initial layers, aiding intricate information extraction, but fewer in later layers, which hampers global semantic understanding. In contrast, coarse-grained pruning preserves more parameters in later layers. To address this, we propose an adaptive algorithm that dynamically fuses coarse-grained and fine-grained importance estimations for each LLM sub-component, automatically adjusting their proportions during learning.

\section{The Proposed Method}
Figure~\ref{fig:suanfa} illustrates our proposed Hybrid-grained Weight Importance Assessment (HyWIA) method, which consists of three distinct steps: the weight grouping step (blue), hybrid-grained assessment step (green), and the fine-tuning step (white). 

\noindent{\textbf{Highlights.}}
Our solution is grounded in the use of Taylor expansion~\cite{NIPS19896c9882bb,molchanov2016pruning} to calculate the fine-grained and coarse-grained gradients derived from the LLM for each input sample.
Subsequently, HyWIA takes these fine-grained and coarse-grained gradients as inputs and performs adaptive fusion based on attention mechanism in an efficient training-free manner.
%
In particular, HyWIA utilizes the attention mechanism to dynamically adjust the importance estimation of fine-grained and coarse-grained metric, such that the model can focus on the most relevant input features, thereby deciding the most suitable assessment for the importance of weights. This dynamic adjustment of weights is based on the input fine-grained and coarse-grained gradients. Consequently, our model can automatically adapt its output results under diverse input conditions, effectively accommodating changes in the input data.

\subsection{Grouping Step} \label{sec:struct}
The first step in pruning involves building groups for LLMs. Assuming $N_{i}$ and $N_{j}$ are two neurons in the model.
The connection between structures can be defined as:
{
\begin{equation}
\label{eq:dependency}
\text{Connect}(N_{i}, N_{j}) = 
\begin{cases} 
w_{ij}, & \\

\sum_{p \in \mathcal{P}(N_{i}, N_{j})} \prod_{(u, v) \in p} w_{uv}, & \\

0, &
\end{cases}
\end{equation}}

\begin{itemize}
    \item $w_{ij}$ if there is a direct connection from $N_{i}$ to $N_{j}$.
    \item $\sum_{p \in \mathcal{P}(N_{i}, N_{j})} \prod_{(u, v) \in p} w_{uv}$ where $\mathcal{P}(N_{i}, N_{j})$ is the set of all paths from $N_{i}$ to $N_{j}$.
    \item 0 if there is no path from $N_{i}$ to $N_{j}$.
\end{itemize}
This formula calculates the connection between neurons $N_{i}$ and $N_{j}$ within the sub-structure, which can be obtained and located through the defined connection relationships. This facilitates the estimation of the importance of each connection structure in LLM in terms of the entirety and the importance of individual elements within the connection structure. Consequently, it aids in the pruning of unimportant connection structures or specific elements within them. The Algorithm~\ref{alg:connection_importance} in the Appendix calculates the importance of connection based on a direct connection, presence of path connection, or no connection.

\begin{algorithm}[tb]
\caption{Attention Fusion Model}
\label{alg:attention_fusion_model}
\textbf{Input}: fine\_grained\_grad, coarse\_grained\_grad \\
\textbf{Parameter}: $d_f$ (dimension of fine-grained gradients), $d_c$ (dimension of coarse-grained gradients), $d_{model}$ (dimension of model) \\
\textbf{Output}: Weight importance score
\begin{algorithmic}[1] 
\STATE Initialize the linear transformations: $W_q$, $W_k$, $W_v$, and $output\_layer$
\STATE Compute $Q = W_q(fine\_grained\_grad)$ 
\STATE Compute $K = W_k(coarse\_grained\_grad)$
\STATE Compute $V = W_v(coarse\_grained\_grad)$ 
\STATE Compute attention weights: $attention\_weights = softmax(\frac{Q \cdot K^T}{\sqrt{d_model}})$
\STATE Compute interaction output: $interaction\_output = attention\_weights \cdot V$
\STATE Compute $\alpha = \text{mean}(attention\_weights, \text{dim}=1)$ \text{ \# Compute mean across attention weights}
\STATE Reshape $\alpha$ to shape $[n, 1]$
\STATE Compute fused output: $fused\_output = \alpha \cdot fine\_grained\_grad + (1 - \alpha) \cdot coarse\_grained\_grad$
\STATE \textbf{return} $fused\_output$
\end{algorithmic}
\end{algorithm}

\subsection{Hybrid-grained Weight Importance Assessment}
\noindent{\textbf{Gradient and importance estimation.}} 
The impact of each parameter on the loss function is estimated by gradients, utilizing the Taylor expansion approximation of the loss deviation function.
Consequently, we utilize this information to estimate the coarse-grained importance and the fine-grained importance.

\noindent{\textbf{Coarse-grained formula.}}
At a coarse level, the pruning mask \( \text{m} \) can be treated as a binary variable where each element indicates whether an entire block, layer, or a group of parameters in the model is kept (1) or pruned (0). The coarse-grained optimization can be represented as:

\begin{equation}
\argmin_{\text{m}_{\text{coarse}}} \loss(\text{m}) 
    \approx \argmin_{\text{m}_{\text{coarse}}} (\mathbb{1} - \text{m}_{\text{coarse}})^\intercal \text{H}_{\text{coarse}} (\mathbb{1} - \text{m}_{\text{coarse}}),
\end{equation}
where \( \text{m}_{\text{coarse}} \) represents the mask at a coarse level, such as entire layers or blocks.

\noindent{\textbf{Fine-grained formula.}}
At a fine-grained level, the mask \( \text{m} \) targets individual neurons, weights, or smaller sub-components of the model. The fine-grained optimization can be represented as:

\begin{equation}
\argmin_{\text{m}_{\text{fine}}} \loss(\text{m}) 
    \approx \argmin_{\text{m}_{\text{fine}}} (\mathbb{1} - \text{m}_{\text{fine}})^\intercal \text{H}_{\text{fine}} (\mathbb{1} - \text{m}_{\text{fine}}),
\end{equation}
where \( \text{m}_{\text{fine}} \) represents the mask at a finer level, such as individual weights or neurons.

\noindent{\textbf{Adaptive fusion.}} 
We propose a dynamic fusion method that combines coarse-grained and fine-grained importance estimations via an adaptive learning network. The complexity of LLMs with multi-layer decoders necessitates both holistic and element-wise assessments, making a single estimation approach insufficient.

Our method adaptively fuses the two criteria through a network that leverages sample-specific loss calculations. This fusion balances computational efficiency and model accuracy, expressed as a weighted combination of coarse- and fine-grained objectives:

\begin{equation}
\begin{aligned}
& \operatorname{argmin}_{\text{m}_{\text{adaptive}}} \loss(\text{m})  \\
    \approx & \operatorname{argmin}_{\text{m}} \alpha  \cdot 
    (\mathbb{1} - \text{m}_{\text{coarse}})^\intercal \mathcal{F}_{\text{coarse}} (\mathbb{1} - \text{m}_{\text{coarse}}) \\
   &  +  (1 - \alpha) \cdot (\mathbb{1} - \text{m}_{\text{fine}})^\intercal \mathcal{F}_{\text{fine}} (\mathbb{1} - \text{m}_{\text{fine}}),
\end{aligned} \label{eq:combine_final_taylor}
\end{equation}
where:
\begin{itemize}
    \item \( \alpha \) is a weighting factor that controls the trade-off between coarse-grained and fine-grained pruning.
    \item \( \mathcal{F}_{\text{coarse}} \) and \( \mathcal{F}_{\text{fine}} \) represent the Hessian's approximations Fisher matrix corresponding to the coarse and fine-grained levels, respectively.
    \item \( \text{m}_{\text{coarse}} \) and \( \text{m}_{\text{fine}} \) are the coarse and fine-grained masks, respectively.
\end{itemize}
\subsection{Algorithm design for adaptive fusion}  \label{sec:adaptive}
\begin{figure*}[t]
  \includegraphics[width=0.48\linewidth]{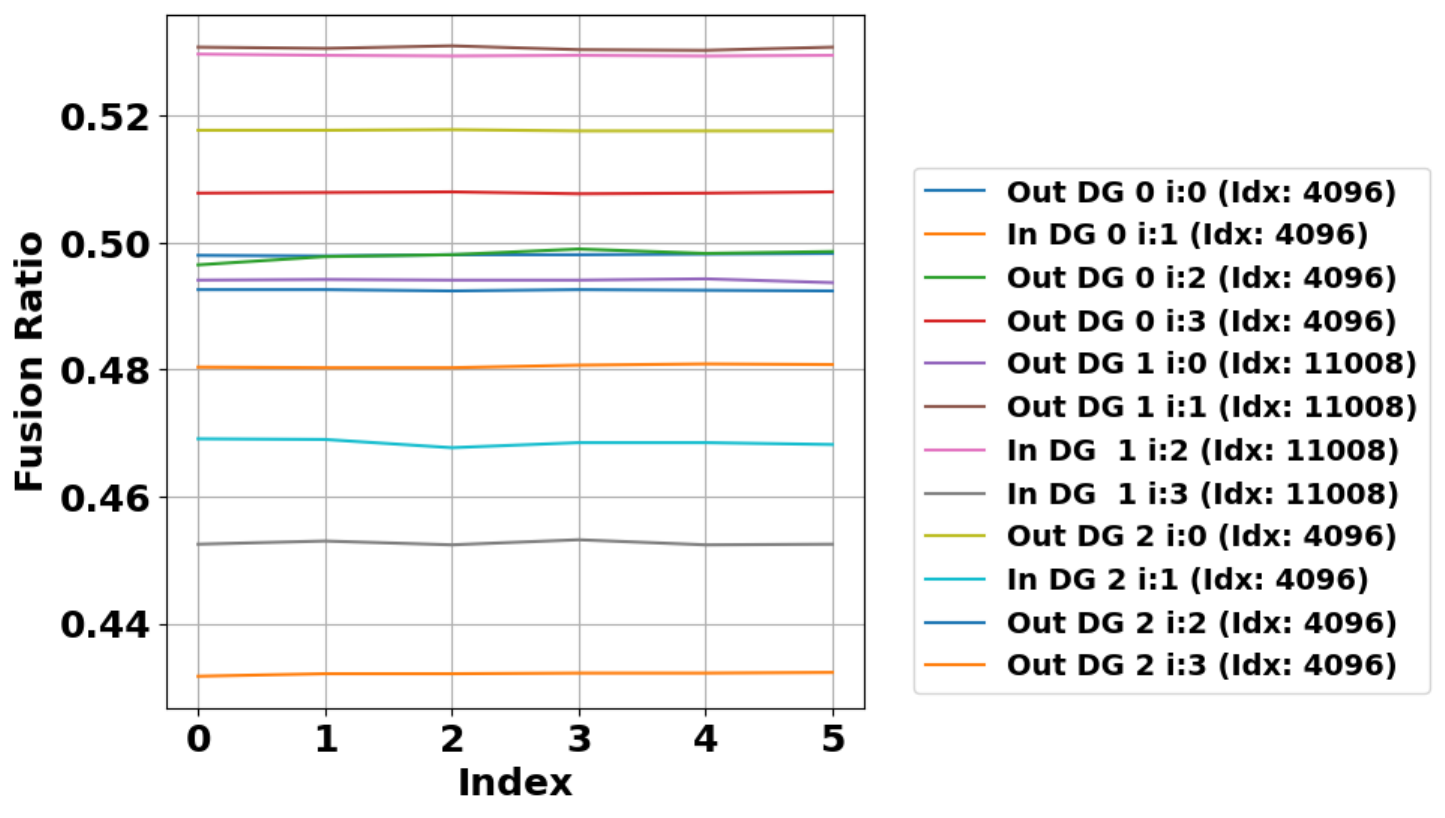} \hfill
  \includegraphics[width=0.48\linewidth]{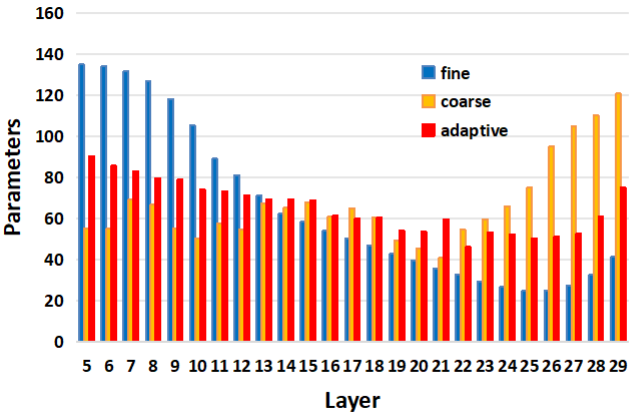}
  \vspace{-0.1cm}
  \caption {On the left, the adaptive fusion rate is shown, where Out DG 1 i:0 (idx:4096) indicates the output channel (Out), direct connection group 1 (DG 1), the 0th sub-group (i:0), and 4096 parameters (idx:4096). For clarity, only the fusion rates of the first six parameters in the first three groups are displayed. On the right, adaptive pruning is compared with fine-grained and coarse-grained methods.}

  \label{fig:rating}
  \vspace{-4mm}
\end{figure*}
To achieve the objective in Eq.\eqref{eq:combine_final_taylor}, we propose the \textit{Attention Fusion Model}, which enables adaptive fusion without traditional parameter training. Algorithm\ref{alg:attention_fusion_model} outlines its workflow, and the key design principles are detailed as follows.

\noindent{\textbf{Dynamic mapping of input features.}}
The algorithm uses three linear transformations, \(W_q\), \(W_k\), and \(W_v\), to map the input \textit{fine\_grained\_grad} and \textit{coarse\_grained\_grad} to a unified dimension (\textit{d\_model}).
Although the parameters of these linear transformations are not updated or trained after model initialization, they still function to map different input features into the same space. Through these mappings, the model can flexibly handle inputs of varying dimensions, thereby adapting to different data characteristics.

\noindent{\textbf{Dynamic weight calculation via attention mechanism.}}
 The attention mechanism computes the dot product between \(Q\) and \(K\) (i.e., \textit{attention\_scores}) to measure the correlation between different features. Then, these correlations are converted into weights (i.e., \textit{attention\_weights}) using the softmax function.
These weights are not fixed; they dynamically change according to different inputs. This means that even though the weight parameters in the model are not trained or updated, the weighted output still adapts dynamically based on the input variations. This dynamic weight allocation is the core manifestation of adaptiveness.

\noindent{\textbf{Flexible fusion of output features.}}
The final interaction output is obtained by calculating the weighted average of 
\(V\), followed by a linear layer to map the output to the desired shape. This attention-based mechanism adaptively fuses different input features, enabling the model to adjust effectively to varying inputs.

\noindent{\textbf{Adaptive fusion without training.}}
Traditional models adjust parameters through training for specific tasks. In contrast, the Attention Fusion Model leverages input characteristics to achieve adaptive fusion via dynamic weight calculation, independent of training data. By utilizing the gradients (fine\_grained\_grad and coarse\_grained\_grad) from LLMs, which inherently carry rich contextual and dynamic features, the model performs adaptive processing through attention mechanisms.

Algorithm~\ref{alg:fine_grained_importance} and Algorithm~\ref{alg:coarse_grained_importance} in Appendix C are prerequisites for implementing Algorithm~\ref{alg:attention_fusion_model}. Each sample is inputted into the LLM to generate gradients, which are connected to the second-order Taylor expansion of the loss function around current weights. Fine-grained estimation accumulates gradients over multiple samples, yielding detailed and accurate parameter importance. In contrast, coarse-grained estimation captures first-order Taylor series information by processing multiple samples simultaneously, providing a direct assessment of each parameter's impact on the loss.

Figure.~\ref{fig:suanfa} illustrates the framework for algorithm implementation, showcasing the interconnection between various modules, and showing the interconnections among different modules. Element-wise multiplication is an operation in which two matrices or tensors of the same dimensions are multiplied together, element by element. 
The details of the estimation methods for importance and element-wise multiplication can be found in Appendix B, Appendix C, and Appendix D.

\noindent{\textbf{Pruning.}} 
Based on the estimation results, the model parameters are sorted according to their respective importance. Subsequently, pruning is performed by removing the importance of these less significant parameters.

\subsection{Fine-tuning Step}
To accelerate the model recovery process and enhance efficiency with constrained data, the low-rank approximation (LoRA)~\citep{hu2022lora} is used to post-train the pruned model. For a pre-trained weight matrix \(m_0 \in \mathbb{R}^{r \times k}\). The update of \(m_0\) is constrained by expressing it through a low-rank decomposition \(m_0 + \Delta m = m_0 + \Gamma \beta\), where \(\Gamma \in \mathbb{R}^{d \times r}\), \(\beta \in \mathbb{R}^{r \times k}\). Throughout training, \(m_0\) remains fixed and does not receive gradient updates, while \(\Gamma\) and \(\beta\) contain trainable parameters. The forward pass is given by:
\begin{equation}
\mathcal{R}(x) = m_0x + \Delta mx = (m_0 + \Gamma \beta) x.
\end{equation}

\section{Experiments}
\subsection{Experimental Setup}
Our experiment is implemented in PyTorch 2.1.2~\cite{paszke2019pytorch}, CUDA 11.6 and HuggingFace 4.29.1~\cite{wolf2019huggingface}, LLaMA-7B-V1/V2, 13B~\cite{touvron2023llama}, Vicuna-7B~\cite{vicuna2023}, BLOOM-7b1~\cite{workshop2022bloom}, Baichuan-7B~\cite{yang2023baichuan},  etc.  All pruning experiments are performed on a single NVIDIA A6000 GPU with 48GB of memory. Benchmark \& Metric can be found in Appendix A.1. Fine-tuning can be found in Appendix A.2. Baselines and configurations can be found in Appendix A.3.

\begin{table*}[tb!]
    \vspace{0em}
     \centering
    \setlength{\tabcolsep}{2.5pt} 
    \small
    \begin{tabular}{c|l|cc|ccccccc|c}
        \hline
        \textbf{Ratio} &  \textbf{Method} & \textbf{WikiT2$\downarrow$} &  \textbf{PTB$\downarrow$} &  \textbf{BoolQ} &  \textbf{PIQA} &  \textbf{HellaS} &  \textbf{WinoG} &  \textbf{ARC-e} &  \textbf{ARC-c} &  \textbf{OBQA} &  \textbf{Ave$\uparrow$} \\
        \hline
        
        \multirow{2}{*}{0\%} & LLaMA-7B~\citep{touvron2023llama} & - & - & 76.5 & 79.8 & 76.1 & 70.1 & 72.8 & 47.6 & 57.2 & 68.59 \\
         & LLaMA-7B$^{\star}$ \citep{ma2023llm} & 12.62 & 22.14 & 73.18 & 78.35 & 72.99 & 67.01 & 67.45 & 41.38 & 42.40 & 63.25 \\
         \hline
        \multirow{9}{*}{20\%} &Magnitude ~\cite{zhang2023loraprune} & 21.78 & 38.64 & 61.89 & 70.81 & 58.34 & 56.87 & 54.87 & 34.02 & 38.40 & 53.59 \\
        &SparseGPT $^{\star}$ \citep{dettmers2023spqr}&- & - &71.13 &75.24	&51.58	&67.56	&68.98	&36.09 	&30.80 	&57.34\\
        &WANDA$^{\star}$ \citep{sun2023simple}& 18.43 & 33.16 & 65.75 & 74.70 & 64.52 & 59.35 & 60.65 & 36.26 & 39.40 & 57.23 \\
        
        & $\text{Element}^2$~\citep{ma2023llm} & 45.70 & 69.33 & 61.47 & 68.82 &  47.56 &  55.09 &  46.46 & 28.24 & 35.20 &  48.98 \\
       
         & LoRAPrune ~\cite{zhang2023loraprune}& {16.80} & \textbf{28.75} & {65.62} & {79.31} & {70.00} & 62.76 & {65.87} & {37.69} & 39.14 &{60.05} \\
        &Compresso ~\cite{guo2023compresso} &- & - & 79.08	 &75.46	&53.44	&67.8	&68.64	&37.97 	&34.20 	&59.51\\
        &FLAP~\cite{an2024fluctuation} &\bf14.62 &- &69.63 &76.82 &\bf71.20 &\bf68.35 &69.91 &39.25 &39.40 &62.08 \\
        &SLEB~\cite{song2024sleb} &18.50 & 31.60 &65.00 &75.00 &65.70 &57.90 &67.06 &36.60 &35.80 &57.60\\
        
       \rowcolor{light-gray} &Ours & 16.42	&31.16 &\bf68.53	&\bf77.8	&70.58	&67.49	&\bf70.24	&\bf40.44 	&\bf42.00 	&\bf62.44\\
        \hline

        \multirow{9}{*}{50\%}&Magnitude ~\cite{zhang2023loraprune} & 78.80 & 164.32 &  47.40 & 54.36 & 33.49 & 53.10 & 37.88 & 26.60 & 30.12 & 40.42 \\
         &SparseGPT $^{\star}$ \citep{dettmers2023spqr}&- & - &64.52	&69.9	&43.29	&64.95	&61.86	&30.37 	&23.80 	&51.24\\

        &WANDA$^{\star}$ \citep{sun2023simple}& 43.89 & 85.87 &  50.90 & 57.38 & 38.12 & 55.98 & 42.68 & 34.20 & 38.78 & 45.43 \\
        
        &$\text{Element}^2$  \cite{ma2023llm} &45.70 &69.33 &61.47 &68.82 &47.56 &55.09 &46.46 &28.24 &35.20 &48.98\\
        &$\text{Vector}$  \citep{ma2023llm} &43.47 &68.51 &62.11 &64.96 &40.52 &51.54 &46.38 &28.33 &32.40 &46.61\\
        &LoRAPrune-8bit ~\cite{zhang2023loraprune}&  33.68 & 53.24 & 61.43 & 70.88 & 47.65 & {55.12} & \textbf{45.78} & 30.50 & 35.62 & 49.56 \\
        & LoRAPrune ~\cite{zhang2023loraprune}& {30.12} & {50.30} & {61.88} & {71.53} & 47.86 & 55.01 & 45.13 & 31.62 & 34.98 & 49.71 \\
       
        &FLAP~\cite{an2024fluctuation}  &31.80& - &60.21 &67.52 &52.14 &\bf57.54 &49.66 &29.95 &35.60 &50.37\\

        \rowcolor{light-gray} &Ours &\bf29.35	&\bf44.38 &\bf60.55	&\bf72.36	&\bf 55.25	&55.09	&\bf50.84	&\bf31.48 	&\bf37.00 &\bf51.80 \\

        \hline
    \end{tabular}
   
    \caption{Zero-shot performance of the compressed LLaMA-7B models. 
    The average is calculated among seven classification datasets. \textbf{Bold} denotes the best performance.
    $^{\star}$ denotes the results obtained by reproduction. 
    } \label{tab:llama_result}
    \vspace{-0cm}
\end{table*}

\subsection{Main Results}
We selected LLaMA-7B as a representative case for analysis. In the scenario with a pruning rate of 20\% and 50\%, Table~\ref{tab:llama_result} presents the comparison results between our method and other methods. In terms of average accuracy, our results stand out as the highest among all methods. Our PPL metric for WikiText2 is the lowest among all methods in the 50\% pruning rate. We also applied our method to Vicuna-7B, Baichuan-7B, Bloom-7B, and LLaMA-7B-V2 yielded identical conclusions.

\begin{table*}[tb!]
    \vspace{0mm}
    \centering
    \setlength{\tabcolsep}{2.5pt} 
    \small
    \begin{tabular}{c|l|cc|ccccccc|c}
         \hline
        \textbf{Ratio} &\textbf{Method} & \textbf{WikiT2$\downarrow$} &  \textbf{PTB$\downarrow$} &  \textbf{BoolQ} &  \textbf{PIQA} &  \textbf{HellaS} &  \textbf{WinoG} &  \textbf{ARC-e} &  \textbf{ARC-c} &  \textbf{OBQA} &  \textbf{Ave$\uparrow$}  \\
        \hline
         \multirow{2}{*}{0\%} 
         &LLaMA-13B~\citep{touvron2023llama} &- &- &78.1	&80.1	&79.2	&73.0 	&74.8	&52.7	&56.4	&70.61 \\
         &LLaMA-13B~\citep{ma2023llm} &  11.58 & 20.24 & 68.47 & 78.89 & 76.24 & 70.09 & 74.58 & 44.54 & 42.00 & 64.97 \\
         \hline
        \multirow{4}{*}{20\%} 
        & L2~\citep{ma2023llm}& 20.97 & 38.05 & 73.25 & 76.77 & 71.86 & 64.64 & 67.59 & 39.93 & 40.80 & 62.12 \\ 
        & Block ~\citep{ma2023llm} & 15.18 & 28.08 & 70.31 & 77.91 & 75.16 &  67.88 & 71.09 &  42.41 &  43.40 & 64.02  \\  
        & FLAP~\cite{an2024fluctuation}  &13.66 &- &72.12 &77.59 &\bf76.01 &\bf69.24 &72.59 &42.56 &43.53 &64.52\\
        \rowcolor{light-gray} &Ours & \bf13.53	&\bf27.55 &\bf72.24	&\bf78.89	&75.63	&67.56	&\bf73.49	&\bf44.11	&42.40	&\bf64.90 \\
        \hline
    \end{tabular}
    \caption{Zero-shot performance of the compressed  LLaMA-13B at 20\% pruning rate.} 
    \label{tbl:13B_result}
    \vspace{-0mm}
\end{table*}

Using the adaptive pruning algorithm, each LLaMA-7B parameter is assigned a fusion ratio for fine- and coarse-grained estimation. Figure~\ref{fig:rating} (left) visualizes the fusion ratios for the first three groups and their initial six sub-group parameters. In the figure, ``out" and ``in"  denote Linear output and input channels, ``DG"  represents the connection group, ``i" is the i-th sub-group, and ``idx'' the parameter count. The fusion ratios within the same channel show minimal differences, while across different dependency groups, they range from 0.4 to 0.6, indicating varying group importance during estimation.

After obtaining this ratio, our algorithm dynamically fuses coarse- and fine-grained estimations to create a comprehensive metric for pruning. The right side of the figure compares the parameter distribution for layers 5-29 of LLAMA-7B pruned by our method with those pruned by fine- and coarse-grained methods. It shows that adaptive pruning balances the importance of both front and back layers, leading to more evenly distributed pruning and optimal results.

In the Appendix, additional illustrations showcasing LLaMA-7B with adaptive pruning can be found in Figure~\ref{sec:fig:line_plot} and Figure~\ref{sec:fig:fuse_ratio}. Details on the number of parameters after pruning for each layer can be accessed in Table~\ref{sec:compare_num}. Hardware cost information is available in Table~\ref{sec:tbl:cost_num}. Comparative analyzes of resource consumption and performance evaluations for the LLaMA-7B, Vicuna-7B, and Bloom-7b1 models are presented in Table~\ref{sec:tbl:Latency_LLaMA}, Table~\ref{sec:tbl:Latency_Vicuna}, and Table~\ref{sec:tbl:Latency_Bloom}. Table~\ref{generate_com} provides generation examples from the original LLaMA-7B and 20\% compressed models. The assessment of computational overhead, including time spent and memory consumption, was conducted using Algorithm~\ref{alg:resource_measurement}. Memory usage of the Adaptive Fusion network on a single NVIDIA A6000 GPU ranged between 1.04 MB and 3.00 MB, with an average processing time of approximately 0.013970 seconds.

\subsection{Ablation Study}
We conducted ablation experiments to analyze the impact of varying sample sizes and pruning rates, systematically assessing performance and the robustness of our approach.\\
\noindent{\textbf{Sample numbers.}}
We experiment with sample numbers 10, 20, 30, 40, and 50 to provide input to the model and compare the impact on accuracy. We investigate whether the sample numbers affect various aspects of training and model performance.
From Table~\ref{tbl:llama_num} and Appendix Table~\ref{sec:tbl:vicuna_num} 
, the first row of each section represents the experimental results for LLM-Pruner $\text{Element}^2$~\cite{ma2023llm}, while the second row displays our experimental results. It is evident that the average accuracy exhibits an increasing trend with the number of example prompts. Concurrently, the perplexity (PPL) of WikiText2 and PTB decreases with increasing sample number. Our model consistently demonstrates higher accuracy compared to LLM-Pruner methods.\\
\noindent{\textbf{Pruning ratio.}}
The choice of pruning rate directly affects the pruning effect and performance of the model. In our experiments, we tried pruning rates of 5\%, 10\%, 20\%, and 50\% to compare the accuracy of the models, studying whether different pruning rates impact the model performance. 
In Table~\ref{tbl:llama_ratio} and Appendix Table~\ref{sec:tbl:vicuna_ratio}, results are categorized into four sections based on pruning rates of 5\%, 10\%, 15\%, and 20\%. The first row in each section shows the experimental outcomes for LLM-Pruner $\text{Element}^2$, while the second row displays our results. Overall, our experimental results outperform the LLM-Pruner method.

We conduct ablation experiments comparing adaptive estimation with coarse-grained estimation and fine-grained estimation in  Appendix Table~\ref{sec:tbl:vicuna_result},  Table~\ref{sec:bloom_result}, \ref{sec:baichuan_result} and \ref{sec:tbl:v2_result}.
We provide ablation experiments with the adaptive algorithm in Appendix Table~\ref{sec:tbl:adap_abl}, Table~\ref{sec:tbl:adap_abl2}, and with the grouping algorithm in Appendix Table~\ref{sec:tbl:llama7B_tune_group}. The performance is analyzed with or without fine-tuning in Appendix Table~\ref{sec:tbl:llama7B_tune} and~\ref{sec:tbl:vica7B_tune}.
\begin{table*}[tb!]
     \vspace{0.1cm}
    \centering
     \setlength{\tabcolsep}{2.5pt} 
    \small
    \begin{tabular}{c|cc|ccccccc|c}
        \hline
       \bf Number &\bf WikiText2$\color{teal}\downarrow$ &\bf PTB$\color{teal}\downarrow$ &\bf BoolQ &\bf PIQA &\bf  HellaSwag &\bf WinoGrande &\bf ARC-e &\bf ARC-c &\bf OBQA & \textbf{Average$\uparrow$} \\
        \hline
        \multirow{2}{*}{10} 
         &17.30  &30.74 &65.14 &76.01 &67.89	&61.4 &51.43 &38.23	&40.6	&57.24\\
         &17.38	&31.16 &67.83 &77.15 &69.81	&65.04	&64.44	&38.74	&41.4	&60.63\\

        \hline
        \multirow{2}{*}{20}  
        &17.28	&31.41  &63.39	&76.28	&68.84	&66.54	&51.98	&37.54	&41.2	&57.96\\
        &17.89	&33.83  &69.14	&77.64	&69.70	&63.46	&64.44	&40.10	&40.80	&60.75\\

         \hline
        \multirow{2}{*}{30} 
        &17.25  &31.41 &63.49	&76.12	&69.04	&66.14	&52.36	&37.80 	&41.20	&58.02\\
	&17.22	&30.93 &67.55	&77.08	&70.15	&65.02	&66.41	&40.27	&41.60 &61.15\\

        \hline
        \multirow{2}{*}{40} 
        & 17.17	&30.68 &67.13 &77.80 &70.02	&62.27	&54.55	&40.27	&40.80	&58.97\\
        & 17.15 &30.66  &68.53	&77.53	&70.30 	&64.96	&68.86	&40.10 	&41.80	&61.73\\

         \hline
        \multirow{2}{*}{50} 
        &17.16 &  30.11 & 64.62 & 77.20 &  68.80 & 63.14 & 64.31 & 36.77 & {39.80} & 59.23 \\
        & 16.42	&31.06 &68.53	&77.8	&70.58	&67.49	&70.24	&40.44 	&42.00 	&62.44\\
     \hline
    \end{tabular}
  
    \caption{Sample numbers for LLaMA-7B at 20\% pruning rate. The first row in each section shows results for LLM-Pruner $\text{Element}^2$~\cite{ma2023llm}.} \label{tbl:llama_num}
     \vspace{0.2cm}
     
\end{table*}\\
\begin{table*}[tb!]
     \vspace{-0.0cm}
    \centering
    
    \setlength{\tabcolsep}{2.6pt} 
    \small
    \begin{tabular}{c|cc|ccccccc|c}
        \hline
        \bf Ratio &\bf WikiText2$\color{teal}\downarrow$ &\bf PTB$\color{teal}\downarrow$ &\bf BoolQ &\bf PIQA &\bf  HellaSwag &\bf WinoGrande &\bf ARC-e &\bf ARC-c &\bf OBQA &\textbf{Average$\uparrow$} \\
        \hline
        \multirow{2}{*}{5\%} 
        &13.01 &23.02 &70.98	&77.78	&72.53	&66.61	&69.48	&42.06	&42.60	&63.14\\
        &12.91 &22.96 &70.60	&77.35	&72.54	&67.01	&70.54	&42.15	&42.20	&63.20\\

       \hline
        \multirow{2}{*}{10\%}  
        &14.02 &24.99 &70.76 &77.62 &71.87	&66.14	&69.73	&42.15	&41.80	&62.86\\
        &14.02 &24.99  &70.54	&78.02	&72.12	&66.43	&70.45	&42.52	&42.20 &63.18\\
         \hline
       
        \multirow{2}{*}{20\%}
        &17.16 &  30.11 & 64.62 & 77.20 &  68.80 & 63.14 & 64.31 & 36.77 & {39.80} & 59.23 \\
        & 16.42	&31.16 &68.53	&77.80	&70.58	&67.49	&70.24	&40.44 	&42.00 	&62.44\\
          \hline
        \multirow{2}{*}{50\%}
        &45.70 &69.33 &61.47 &68.82 &47.56 &55.09 &46.46 &28.24 &35.20 &48.98\\
        &29.35	&44.38 &60.55	&72.36	& 55.25	&55.09	&50.84	&31.48 	&37.00 &51.80 \\

        
    \hline
    \end{tabular}
  
   \caption{Prune ratio for LLaMA-7B with 50 samples. The first row in each section shows results for LLM-Pruner $\text{Element}^2$~\cite{ma2023llm}.} \label{tbl:llama_ratio}
    \vspace{-0.4cm}
\end{table*}
From Table~\ref {tbl:llama_ratio} and Appendix Table~\ref{sec:tbl:vicuna_ratio}, it can be observed that our experimental results overall outperform the fine-grained method. With increasing pruning rates, parameters, MACs, memory, and latency consistently decrease.

\section{Related Work}
\noindent{\textbf{Pruning for LLMs.}}
Various pruning techniques \cite{li2022pruning,yang2023pruning,wu2022compiler,kong2022spvit,shen2024search} have been developed to reduce the model size and inference cost.
\textit{PtPF}~\cite{kwon2022fast} proposes a fast post-training pruning framework for Transformers, eliminating the need for retraining
\textit{FGlP}~\cite{lee2020flexible} employs group-level pruning to accelerate deep neural networks.
\textit{CoFi}~\cite{xia2022structured} prunes both coarse-grained and fine-grained modules by using masks of varying granularity to control the pruning of each parameter.
\textit{LoRAPrune}~\cite{zhang2023loraprune} designed a LoRA-guided pruning criterion, which uses the weights and gradients of LoRA.
\textit{FLAP}~\cite{an2024fluctuation} developed structured importance indicators, and the adaptive search globally compresses the model. 
\textit{COMPRESSO}~\citep{guo2023compresso} introduced a collaborative prompt that promotes collaboration between the LLM and the pruning algorithm.
\textit{PAP}~\cite{zhangplug} proposed a pruning metric that effectively combines weight and activation information in LLM, 
\textit{SLEB}~\cite{song2024sleb} is devised to optimize LLMs through the removal of redundant transformer blocks. 
\textit{Shortened LLaMA}~\cite{kim2024mefomo} enhances inference speeds, particularly in memory-constrained scenarios with limited batch sizes for LLM execution.
\textit{CompactPKD}~\cite{muralidharan2024compact} integrating depth, width, attention, and MLP pruning, along with knowledge distillation-driven retraining.
\textit{Bonsai}~\cite{dery2024everybody} devise a perturbative pruning approach devoid of gradients, capable of producing compact, swift, and precise pruned models.

\noindent{\textbf{Efficient learning for LLMs.}}
The goal of efficient learning~\cite{liu2024tsla,liu2024brain,liu2021explainable,liu2022efficient,liu2023scalable,liu2023interpretable,li2020efficient,zhan2024fast} is to achieve better results with fewer resources. \textit{SpQR} \cite{dettmers2023spqr} employed a method involving the identification and isolation of outlier weights. 
\textit{LLM-FP4}~\cite{liu2023llmfp} suggests FP4 as a post-training method to quantify weights and activations in large language models (LLM) up to floating point values of 4 bits. 
\textit{QLORA} \cite{dettmers2023qlora} introduces methods to save memory, which is information-theoretically optimal for normally distributed weights. 
\textit{Less} \cite{liang2023less} proposes Task-aware layer-wise distillation (TED) as a solution to reducing the knowledge gap between teacher and student models.
\textit{MiniLLM}~\cite{gu2023knowledge} put forth a knowledge distillation approach aimed at condensing LLMs into more compact language models.
\textit{LoRD}~\cite{kaushal2023lord} utilizes Low Rank Decomposition (LoRD) to ensure that the compressed model remains compatible with the cutting-edge near-lossless quantization method.
\textit{RoRA}~\cite{liu2025rora} proposes Rank-adaptive Reliability Optimization (RoRA), which optimizes LoRA's scaling factor by replacing $\alpha/r$ with $\alpha/\sqrt{r}$, ensuring improved performance.
\textit{AdaPTwin}~\cite{biju2024adaptwin} compresses pairs of weight matrices that are dependent on products within the transformer attention layer simultaneously.
\section{Conclusion }
In this paper, we observe that coarse-grained and fine-grained pruning generate different sparsity distributions across LLM layers. We suggest that evaluating both holistic and individual assessments of weight importance is essential for LLM pruning. We introduce Hybrid-grained Weight Importance Assessment (HyWIA), a novel method that merges fine-grained and coarse-grained evaluations of weight importance for pruning LLMs. Leveraging an attention mechanism, HyWIA adaptively determines the optimal blend of granularity in weight importance assessments in an end-to-end pruning manner. Experiments on LLaMA-V1/V2, Vicuna, Baichuan, and Bloom across various benchmarks demonstrate HyWIA's effectiveness in pruning LLMs.

\bibliography{aaai25}
\clearpage

\newpage
\clearpage
\onecolumn
\appendix

\section{Appendix}

\section{A \quad Detailed Experimental Settings} \label{app:exp:set}
\subsection{A.1 \quad Benchmark \& Metric} \label{sec:bench}
The model was evaluated on datasets covering a range of natural language understanding and reasoning challenges, including common sense reasoning, physical interaction understanding, and coreference resolution, using the EleutherAI LM Harness~\cite{eval-harness}\footnote{https://github.com/EleutherAI/lm-evaluation-harness}. BoolQ~\cite{clark2019boolq} assesses the model's accuracy in providing correct answers to questions. PIQA~\cite{Bisk2020piqa} evaluates the model's performance using accuracy related to question answering. HellaSwag~\cite{zellers2019hellaswag} assess the model's ability to correctly predict endings. WinoGrande~\cite{ai2:winogrande} assesses the model's understanding of gender-related information, potentially using accuracy and other indicators. Arc Easy~\cite{allenai:arc} and Arc Challenge~\cite{allenai:arc} evaluate the model's performance in answering common-sense reasoning questions. WikiText2~\cite{merity2016pointer} focuses on predicting the next word in a sequence, PTB~\cite{marcus-etal-1993-building} focuses on syntactic parsing and understanding grammatical relationships within sentences. Perplexity (PPL) is used to measure the predictive capability of a language model on a given text sequence.

\subsection{A.2 \quad Fine-tuning} \label{sec:LoRA}
We employ popular Parameter-Efficient Fine-tuning (PEFT)~\cite{peft} methodologies, leveraging Half-precision floating-point (fp16) for fine-tuning our pruned LLMs generated by the adaptive fusion method.  The fine-tuning dataset is obtained from yahma/alpaca-cleaned, and we utilize the Adam optimizer with a learning rate of $1\times 10^{-4}$. During the fine-tuning phase, we set the LoRA rank to 8 and $\alpha$ to 16, employing a batch size of 64. Fine-tuning on the NVIDIA A6000 GPU typically requires only 3 to 4 hours to complete. The hyperparameters described in LoRA~\citep{hu2022lora}.

\subsection{A.3\quad Baseline and Configurations} 
The baseline provides unpruned test results, demonstrating the performance metrics of LLaMA-7B with 50 samples, pruning from the 4th layer to the 29th layer. Specific performance metrics include estimation scores for BoolQ~\cite{clark2019boolq}, PIQA~\cite{Bisk2020piqa}, HellaSwag~\cite{zellers2019hellaswag}, WinoGrande~\cite{ai2:winogrande}, ARC-e~\cite{allenai:arc}, ARC-c~\cite{allenai:arc}, OBQA ~\cite{OpenBookQA2018} tasks, as well as the average accuracy and Perplexity (PPL) for WikiText2~\cite{merity2016pointer} and PTB~\cite{marcus-etal-1993-building}. Perplexity is used to measure the predictive capability of a language model in a given text sequence. 
\section{B\quad Algorithm for Grouping  } 
The Algorithm~\ref{alg:connection_importance} calculates the connection importance between neurons \( N_{i} \) and \( N_{j} \) within the sub-group, which aids in estimating the importance of various connection structures within large language models and assists in pruning unimportant connection structures or specific elements. This algorithm helps in determining how crucial each connection between neurons is by considering both direct connections and indirect paths. \\
The algorithm's ability to handle multiple paths ($\mathcal{P}(N_{i}, N_{j})$) between neurons means it can be adapted to various network architectures and connection patterns.

\begin{algorithm}[tb]
\caption{Grouping Algorithm}
\label{alg:connection_importance}
\textbf{Input}: Set of neurons $\mathcal{N}$, Connection weights $w_{uv}$\\
\textbf{Output}: Connection importance between neurons $N_i$ and $N_j$
\begin{algorithmic}[1] 
\FOR{each pair of neurons $(N_i, N_j)$ in $\mathcal{N}$}
    \IF{there is a direct connection from $N_i$ to $N_j$}
        \STATE $\text{Connect}(N_{i}, N_{j}) \gets w_{ij}$
    \ELSIF{there exists at least one path from $N_i$ to $N_j$}
        \STATE $\text{Connect}(N_{i}, N_{j}) \gets \sum_{p \in \mathcal{P}(N_{i}, N_{j})} \prod_{(u, v) \in p} w_{uv}$
    \ELSE
        \STATE $\text{Connect}(N_{i}, N_{j}) \gets 0$
    \ENDIF
\ENDFOR
\STATE \textbf{return} $\text{Connect}(N_{i}, N_{j})$
\end{algorithmic}
\end{algorithm}

\section{C\quad  Algorithm for Importance Estimation } 
\subsection{C.1 \quad  Fine-Grained Estimation of Importance } \label{sec:fineest}
We provide Algorithm~\ref{alg:fine_grained_importance} for fine-grained estimation of importance. This algorithm uses the gradients of the loss function to estimate the importance of each parameter in a neural network model. By accumulating gradients over batches and calculating the Fisher Information Matrix, it provides a fine-grained estimation of parameter importance, which can be useful for tasks like pruning less important parameters to reduce the model's size while maintaining performance.

\begin{algorithm}[tb]
\caption{Fine-Grained Estimation of Importance Using Taylor Series} 
\label{alg:fine_grained_importance}
\textbf{Input}: Neural network model $model$, DataLoader $data\_loader$, Loss function $loss\_fn$, Initial mask $initial\_mask$\\
\textbf{Output}: Importance scores $importance\_scores$
\begin{algorithmic}[1] 
\STATE Set $model$ to evaluation mode.
\STATE Initialize an empty dictionary $gradients$.
\FOR{each batch $(X, y)$ in $data\_loader$}
    \STATE Perform a forward pass to compute the output $output = model(X)$.
    \STATE Compute the loss $loss = loss\_fn(output, y)$.
    \STATE Perform a backward pass to compute gradients $\nabla_{\text{m}} \mathcal{L}$ with respect to parameters.
    \FOR{each parameter $param$ with name $name$ in $model$}
        \IF{$param$ requires gradient}
            \IF{$name$ not in $gradients$}
                \STATE Initialize $gradients[name]$ with $param.grad.clone()$.
            \ELSE
                \STATE Accumulate $gradients[name] \mathrel{+}= param.grad.clone()$.
            \ENDIF
        \ENDIF
    \ENDFOR
\ENDFOR
\STATE Initialize an empty dictionary $fisher\_information$.
\FOR{each $name, grad$ in $gradients$}
    \STATE Compute Fisher Information Matrix approximation $fisher\_information[name] = \text{mean}(grad^2)$.
\ENDFOR
\STATE Initialize an empty dictionary $importance\_scores$.
\FOR{each $name, fisher$ in $fisher\_information$}
    \STATE Set $importance\_scores[name] = fisher$.
\ENDFOR
\STATE \textbf{return} $importance\_scores$
\end{algorithmic}
\end{algorithm}

\subsection{C.2 \quad  Coarse-Grained Estimation of Importance } \label{sec:coarseest}
We provide Algorithm~\ref{alg:coarse_grained_importance} for coarse-grained estimation of importance. This algorithm essentially helps in determining which parts of the model are most crucial by computing their importance based on the gradients of the loss function. These scores can then be used for pruning less important components, potentially improving model efficiency while preserving performance.
\begin{algorithm}[tb]
\caption{Coarse-Grained Estimation of Importance Using Taylor Series}  \label{alg:coarserad}
\label{alg:coarse_grained_importance}
\textbf{Input}: Neural network model $model$, DataLoader $data\_loader$, Loss function $loss\_fn$, Initial mask $initial\_mask$\\
\textbf{Output}: Importance scores $importance\_scores$
\begin{algorithmic}[1] 
\STATE Set $model$ to evaluation mode.
\STATE Initialize an empty dictionary $coarse\_gradients$.
\FOR{each batch $(X, y)$ in $data\_loader$}
    \STATE Perform a forward pass to compute the output $output = model(X)$.
    \STATE Compute the loss $loss = loss\_fn(output, y)$.
    \STATE Perform a backward pass to compute gradients $\nabla_{\text{m}} \mathcal{L}$ with respect to coarse-grained components.
    \FOR{each component (e.g., layer or block) $component$ in $model$}
        \STATE Compute the gradient $grad$ of $loss$ with respect to $component$.
        \IF{$component$ not in $coarse\_gradients$}
            \STATE Initialize $coarse\_gradients[component]$ with $grad.clone()$.
        \ELSE
            \STATE Accumulate $coarse\_gradients[component] \mathrel{+}= grad.clone()$.
        \ENDIF
    \ENDFOR
\ENDFOR
\STATE Initialize an empty dictionary $fisher\_information$.
\FOR{each $component, grad$ in $coarse\_gradients$}
    \STATE Compute Fisher Information Matrix approximation $fisher\_information[component] = \text{mean}(grad^2)$.
\ENDFOR
\STATE Initialize an empty dictionary $importance\_scores$.
\FOR{each $component, fisher$ in $fisher\_information$}
    \STATE Set $importance\_scores[component] = fisher$.
\ENDFOR
\STATE \textbf{return} $importance\_scores$
\end{algorithmic}
\end{algorithm}

\section{D \quad Element-Wise Multiplication} \label{sec:multi}
Assume:
\begin{itemize}
    \item $fine\_grained\_grad$ is a vector $\mathbf{a}$
    \item $coarse\_grained\_grad$ is a vector $\mathbf{b}$
    \item $ratio\_weight$ represents the proportional weight of fine-grained and coarse-grained metrics and is denoted as a vector $\mathbf{w}$.
\end{itemize}
The computation process is as follows:
First, we need to perform element-wise multiplication between {w} and {a}. Element-wise multiplication means that each element of the vectors is multiplied individually, and the result is also a vector.
\begin{align*}
\mathbf{w} \odot \mathbf{a} &= 
\begin{bmatrix} 
w_1 \\ 
w_2 \\ 
\vdots \\ 
w_n 
\end{bmatrix} 
\odot 
\begin{bmatrix} 
a_1 \\ 
a_2 \\ 
\vdots \\ 
a_n 
\end{bmatrix} 
= 
\begin{bmatrix} 
w_1 a_1 \\ 
w_2 a_2 \\ 
\vdots \\ 
w_n a_n 
\end{bmatrix}
\end{align*}

Second, we need to perform element-wise multiplication between the vector (1 - {w}) and {b}. Similar to the previous step, this operation is performed element by element, resulting in a new vector.
{\small
\begin{align*}
(1 - \mathbf{w}) \odot \mathbf{b} &= 
\begin{bmatrix} 
1 - w_1 \\ 
1 - w_2 \\ 
\vdots \\ 
1 - w_n 
\end{bmatrix} 
\odot 
\begin{bmatrix} 
b_1 \\ 
b_2 \\ 
\vdots \\ 
b_n 
\end{bmatrix} 
= 
\begin{bmatrix} 
(1 - w_1) b_1 \\ 
(1 - w_2) b_2 \\ 
\vdots \\ 
(1 - w_n) b_n 
\end{bmatrix}
\end{align*}}

Finally, to get the final estimation output, we sum the results of the two element-wise multiplications performed in the previous steps:
\begin{align*}
\text{estimation\_output} &= \mathbf{w} \odot \mathbf{a} + (1 - \mathbf{w}) \odot \mathbf{b}
\end{align*}

\section{E \quad More Figures on LLaMA-7B}
\begin{figure}[t]
\centering
  \includegraphics[width=.6\columnwidth]{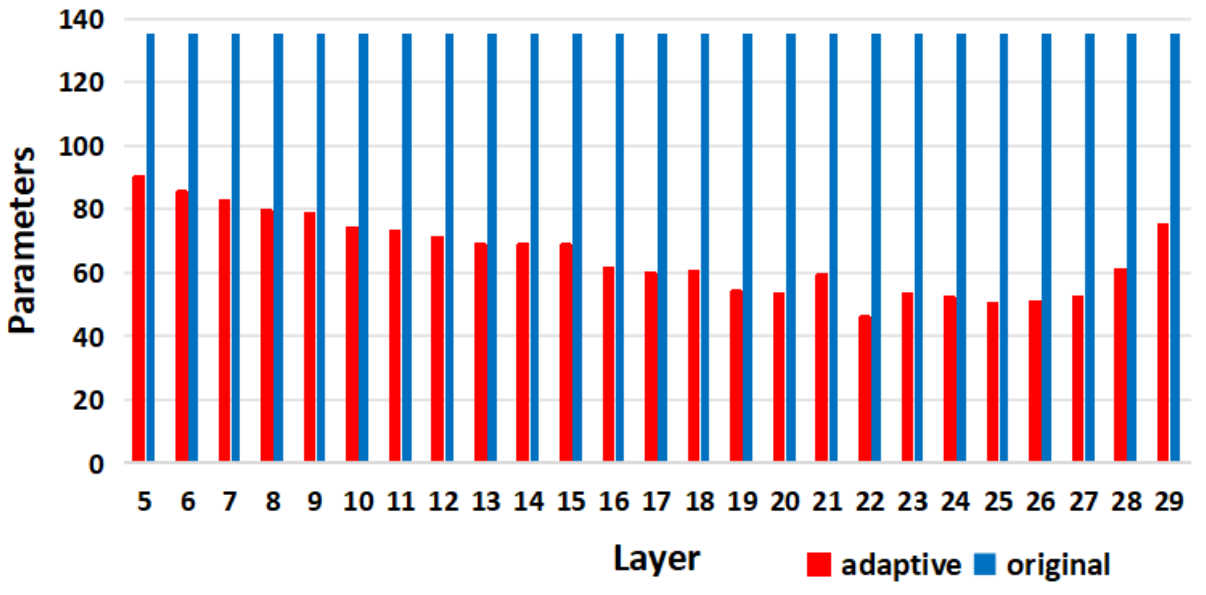} \hfill
  
  \caption {Comparison of LLaMA-7B layer parameters before and after 50\% pruning using our method.}
  \label{sec:fig:compare_plot}
\end{figure}

\begin{figure}[t]
\centering
  \includegraphics[width=.5\linewidth]{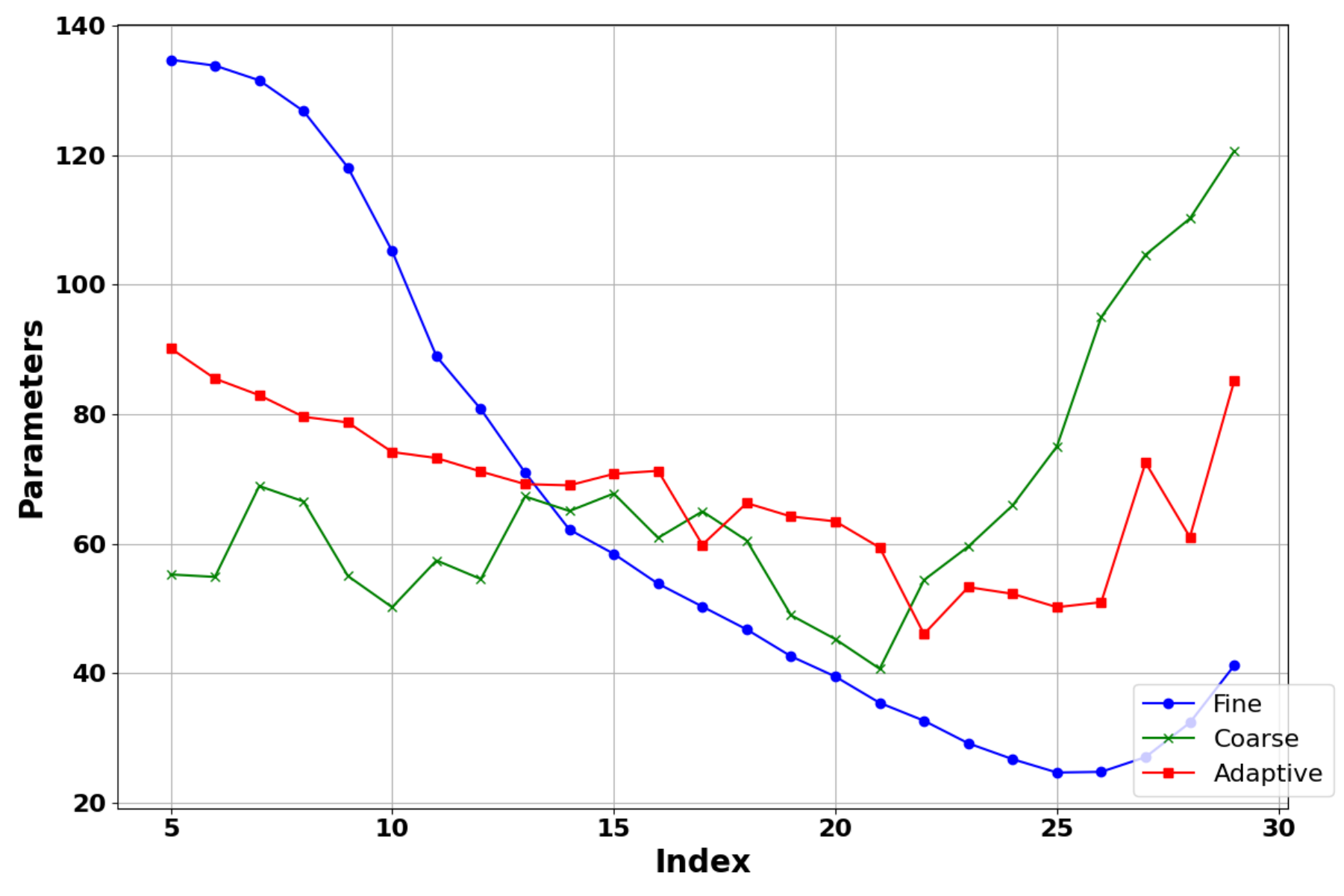} \hfill
 
  \caption {A line plot compares LLaMA-7B with adaptive pruning against fine-grained and coarse-grained methods, all with a 50\% pruning rate.}
  \label{sec:fig:line_plot}
\end{figure}

\begin{figure*}[t]
  \includegraphics[width=0.9\linewidth]{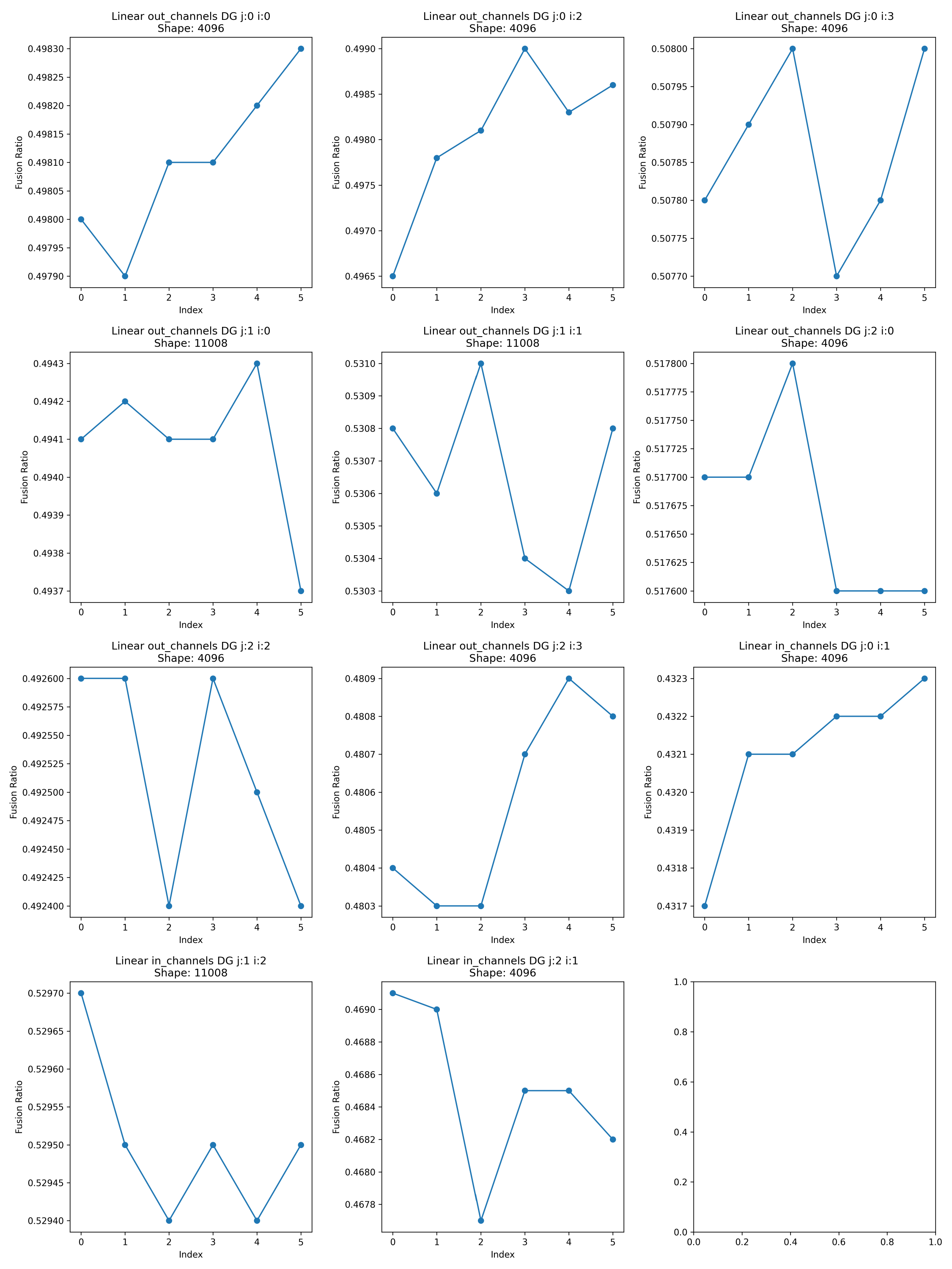} \hfill
 
  \caption {A figure of a line plot showcasing the fusion rate of parameters within each channel in different groups.}
  \label{sec:fig:fuse_ratio}
\end{figure*}

In Figure~\ref{sec:fig:compare_plot}, we compared the parameters of each layer after pruning LLaMA-7B using our method and before pruning. In Figure~\ref {sec:fig:line_plot}, a comparison is made between adaptive pruning and the pruning methods of fine-grained estimation and coarse-grained estimation according to Table~\ref{sec:compare_num}. After applying our pruning method, the parameter distribution across different layers of the pruned model becomes more uniform. In Figure~\ref{sec:fig:fuse_ratio}, presents the fusion rate of parameters within each channel across different groups. With our approach, each parameter is assigned an individual fusion ratio.

\section{F \quad Hardware Cost}
In Table~\ref{sec:tbl:cost_num}, we present the hardware cost calculation for LLaMA-7B when using a pruning rate of 20\%. Through the implementation of our pruning method, we effectively minimize the number of model parameters and reduce memory requirements, resulting in optimized hardware utilization.

\begin{table}[h]
    \vspace{-0mm}
    \centering
    \resizebox{0.6\linewidth}{!}{ 
    \small 
    \begin{tabular}{l|l|ccc}
        \hline
        \bf Ratio &\bf Method &\bf Params &\bf Memory &\bf MAC\\ \midrule
        0\%  &LLaMA-7B &6.74B &1284.5MiB &424.02G\\
        \hline
        \multirow{4}{*}{\parbox{1.0cm}{20\%}} &Wanda \citep{sun2023simple}   &6.74B &12916.5MiB &-\\
             &LLM-Pruner Block ~\citep{ma2023llm}   &5.42B &10375.5MiB &339.60G\\
             &FLAP ~\cite{an2024fluctuation}    &5.07B &9726.2MiB  &- \\
             &Ours     &\bf 4.97B &\bf 9555.8MiB &\bf 312.23G\\
     \hline
    \end{tabular}
    }
   \caption{Hardware Cost for LLaMA-7B with a pruning rate=20\%.} \label{sec:tbl:cost_num}
\end{table}

\section{G \quad Resource consumption and performance evaluation}
In Table~\ref{sec:tbl:Latency_LLaMA}, Table~\ref{sec:tbl:Latency_Vicuna}, and Table~\ref{sec:tbl:Latency_Bloom}, we present a comparison of adaptive fusion method with coarse-grained Latency on the LLaMA-7B, Vicuna-7B, and Bloom-7b1 models using NVIDIA A6000.
\begin{table}[h]
    \vspace{-0mm}
    \centering
    \resizebox{0.7\linewidth}{!}{ 
    \small 
    \begin{tabular}{l|l|cccc}
        \hline
        \bf Ratio &\bf Method  &\bf Params &\bf Memory &\bf MAC &Latency\\ \midrule
       
        0\%  &LLaMA-7B\cite{touvron2023llama} &6.74B &12884.5MiB &424.02G &69.16s\\
        \hline
        \multirow{4}{*}{\parbox{1.0cm}{20\%}} &LLM-Pruner Vector \citep{ma2023llm}  &5.38B &10926.8MiB &328.82G &47.56s\\
             &$\text{LLM-Pruner Element}^2$ ~\cite{ma2023llm}   &5.42B &10375.5MiB &339.60G &43.23s\\
             &Ours    &\bf4.97B &\bf9555.8MiB &\bf312.23G &\bf42.41s\\
       
    \hline
    \end{tabular}
    }
   \caption{Latency for LLaMA-7B with a pruning rate=20\%. The $\text{LLM-Pruner Vector}$~\citep{ma2023llm} is a coarse-grained method, $\text{LLM-Pruner Element}^2$~\citep{ma2023llm}} \label{sec:tbl:Latency_LLaMA}
\end{table}

\begin{table}[h]
    \vspace{-0mm}
    \centering
    \resizebox{0.7\linewidth}{!}{ 
    \small 
    \begin{tabular}{l|l|cccc}
        \hline
        \bf Ratio &\bf Method  &\bf Params &\bf Memory &\bf MAC &\bf Latency\\  \hline
        0\%  &Vicuna-7B~\cite{vicuna2023} &6.73B &425.12G &12924.65MiB & 72.54s\\
        \hline
        \multirow{4}{*}{\parbox{1.0cm}{20\%}} &LLM-Pruner Vector \citep{ma2023llm} &5.71B  &358.82G & 10958.80MiB  &44.68s\\
             &$\text{LLM-Pruner Element}^2$ ~\cite{ma2023llm} &5.53B &347.36G & 10837.12MiB   &44.81s\\
             &Ours    &\bf 5.36B  &\bf 339.70G &\bf 10796.33MiB &\bf43.11s\\
       
    \hline
    \end{tabular}
    }
  
   \caption{Latency for Vicuna-7B with a pruning rate=20\%.}  \label{sec:tbl:Latency_Vicuna} 
\end{table}

\begin{table}[h]
    \vspace{-0mm}
    \centering
    \resizebox{0.7\linewidth}{!}{ 
    \small 
    \begin{tabular}{c|l|cccc}
        \hline
        \bf Ratio &\bf Method  &\bf Params &\bf Memory &\bf MAC &\bf Latency\\   \hline 
        0\%  &Bloom-7b1~\cite{workshop2022bloom} & 7.00B  & 452.91G & 13491.20MiB  & 66.89s\\
       \hline
        \multirow{3}{*}{20\%} &LLM-Pruner Vector \citep{ma2023llm} & 5.68B  &369.03G & 10994.85MiB  &50.49s\\
             &$\text{LLM-Pruner Element}^2$ ~\cite{ma2023llm}   & 5.54B &357.13G & 10972.30MiB &53.41s\\
             &Ours    &\bf 5.38B &\bf 355.27G &\bf 10788.21MiB  &\bf 51.79s\\
       
    \hline
    \end{tabular}
    }
   \caption{Latency for Bloom-7b1 with a pruning rate=20\%. } \label{sec:tbl:Latency_Bloom}
\end{table}

\section{H \quad Ablation Study for Adaptive Fusion Estimation }  \label{sec:adap_fusion} 
\subsection{H.1 \quad Compare Adaptive Fusion Estimation with No Fusion Estimation} 
We employ the same adaptive estimation methodology to evaluate Vicuna-7B with a pruning rate of 20\%, Baichuan-7B, Bloom-7b1 with a pruning rate of 25\%, and LLaMA-7B-V2 with a pruning rate of 50\%. The evaluation results for each model can be found in Table~\ref{sec:tbl:vicuna_result},  Table~\ref{sec:bloom_result}, Table~\ref{sec:baichuan_result}, Table~\ref{sec:tbl:v2_result}. According to the definition of fine-grained ~\cite{xia2022structured} and coarse-grained~\cite{lee2020flexible}, the $\text{LLM-Pruner Vector}$~\citep{ma2023llm} is a coarse-grained method, $\text{LLM-Pruner Element}^2$~\citep{ma2023llm}, is a fine-grained method.  $^{\star}$ denotes the results obtained by reproduction.\\

\begin{table*}[h]
    \vspace{2mm}
    \centering
    
    \resizebox{\linewidth}{!}{
    \begin{tabular}{c|l|cc|ccccccc|c}
        \hline
        \textbf{Ratio} &\textbf{Metho}d & \textbf{WikiText2$\color{teal}\downarrow$} &  \textbf{PTB$\downarrow$} &  \textbf{BoolQ} &  \textbf{PIQA} &  \textbf{HellaSwag} &  \textbf{WinoGrande} &  \textbf{ARC-e} &  \textbf{ARC-c} &  \textbf{OBQA} &  \textbf{Average$\uparrow$}   \\
        \hline
        \multirow{1}{*}{0\%}
        &Vicuna-7B~\cite{vicuna2023} &16.23  &58.19  &75.66	&77.80	&71.05	&67.64	&65.02	&39.93	&42.20	&62.76 \\ 
        \hline
 
       \multirow{3}{*}{20\%}
        &LLM-Pruner Vector~\citep{ma2023llm} &19.94 & \bf 74.66 & 63.15 & 74.59 & 61.95 & 60.30 & 60.48 & 36.60 & \bf 39.40 & 56.64 \\
        &$\text{LLM-Pruner Element}^2$ ~\citep{ma2023llm}& 
        \bf 18.97 & 76.78 & 60.40 & 75.63 & \bf 65.45 & \bf 63.22 & \bf 63.05 & \bf 37.71 & 39.00 & \bf 57.78 \\
        \rowcolor{light-gray}  &Ours &\bf 16.63	&\bf 29.61	&\bf72.78	&\bf 77.33	&\bf68.74	&\bf66.56	&\bf64.29	&\bf38.14	&\bf42.00	&\bf61.35
  \\
    \hline
    \end{tabular}
   }
   \caption{Zero-shot Performance of the compressed  Vicuna-7B with a pruning rate=20\%.} 
    \label{sec:tbl:vicuna_result} 
\vspace{1mm}
\end{table*}
\begin{table*}[h]
    \centering
   
    \resizebox{\linewidth}{!}{
    \begin{tabular}{c|l|cc|ccccccc|c}
        \hline
        \textbf{Ratio} & \bf Method &\bf WikiText2$\color{teal}\downarrow$ &\bf PTB$\color{teal}\downarrow$ & \bf BoolQ & \bf PIQA & \bf HellaSwag & \bf WinoGrande & \bf ARC-e & \bf ARC-c & \bf OBQA & \textbf{Average$\uparrow$}   \\
        \hline
        \multirow{1}{*}{0\%}
        &Bloom-7b1~\cite{workshop2022bloom} &-  &-  & 62.91 & 73.56 & 59.67 & 64.40 & 57.28 & 33.53 & 36.00 & 55.34  \\
        \hline
        \multirow{3}{*}{25\%}
       & LLM-Pruner Vector$^{\star}$~\citep{ma2023llm} & 101.20 & 319.13 & 61.19 & 71.16 & 47.65 & 55.56 & 50.38 & 30.89 & 32.8 & 49.95   \\
       & $\text{LLM-Pruner Element}^2$$^{\star}$~\citep{ma2023llm} & \bf 101.20 &\bf 319.13 & 61.62 & 70.40 & 48.28 & 56.12 & 50.42 & 30.12 & 34.4 & 50.19  \\
               
        \rowcolor{light-gray} &Ours & 197.38 & 586.98 & \bf 62.33	&\bf 71.16	&\bf 49.49	&\bf 57.73	&\bf 52.10 	& \bf 31.14	&\bf 35.8	&\bf 51.39  \\
    \hline
    \end{tabular}
   }
    \caption{Zero-shot Performance of the compressed Bloom-7b1 with a pruning rate=25\%. } 
    \label{sec:bloom_result}
\end{table*}
\begin{table*}[h]
 
    \centering
    
    \resizebox{\linewidth}{!}{
    \begin{tabular}{c|l|cc|ccccccc|c}
        \hline
       \textbf{Ratio} & \bf Method & \bf WikiText2$\color{teal}\downarrow$ & \bf PTB$\color{teal}\downarrow$ & \bf BoolQ & \bf PIQA & \bf HellaSwag &\bf  WinoGrande & \bf ARC-e & \bf ARC-c & \bf OBQA &\textbf{Average$\uparrow$} \\
       \hline
       \multirow{1}{*}{0\%}
        &Baichuan-7b~\cite{yang2023baichuan} &- & - &68.35	&76.39	&67.18	&62.98	&56.05	&38.14	&42.8	&58.93  \\
        \hline
        \multirow{3}{*}{25\%}
        &LLM-Pruner Vector$^{\star}$~\citep{ma2023llm} &25.74 &90.72 &61.47	&74.59	&61.54	&61.40	&50.00	&33.87	&38.40	&54.47   \\
        &$\text{LLM-Pruner Element}^2$$^{\star}$~\citep{ma2023llm} &20.96 &81.96 &61.62	&73.07	&59.87	&54.70	&49.92	&33.45	&37.80	&52.92  \\
        
        \rowcolor{light-gray} &Ours &\bf 20.68 &\bf 81.00 &\bf 63.05	&\bf 75.38	&\bf 62.93	&\bf 57.83	&\bf 51.04	&\bf 34.76	&\bf 39.6	&\bf 54.94  \\
    \hline
    \end{tabular}
   }
   \caption{Zero-shot Performance of the compressed Baichuan-7B with a pruning rate=25\%. } 
    \label{sec:baichuan_result}
\end{table*}
\begin{table*}[h]
    \vspace{1mm}
    \centering
    
    \resizebox{\linewidth}{!}{
    \begin{tabular}{c|l|cc|ccccccc|c}
        \hline
        \textbf{Ratio} &\textbf{Metho}d & \textbf{WikiText2$\downarrow$} &  \textbf{PTB$\downarrow$} &  \textbf{BoolQ} &  \textbf{PIQA} &  \textbf{HellaSwag} &  \textbf{WinoGrande} &  \textbf{ARC-e} &  \textbf{ARC-c} &  \textbf{OBQA} &  \textbf{Average$\uparrow$}   \\
        \hline
        \multirow{1}{*}{0\%}
        &LLaMA-2-7B~\cite{vicuna2023} &13.99	&28.99  &42.97	&76.06	&70.02	&65.51	&63.05	&36.52	&40.80 	&56.42 \\ 
        \hline
 
        \multirow{3}{*}{50\%}
        &LLM-Pruner Vector$^{\star}$~\citep{ma2023llm} & 90.01	&214.25	&38.26	&69.53	&47.98	&52.49	&48.44	&28.16	&36.00 	&45.84\\
       
        &$\text{LLM-Pruner Element}^2$$^{\star}$ ~\citep{ma2023llm}& 99.63	&258.44	&37.21	&68.77	&49.31	&51.28	&46.51	&28.50 	&35.20 	&45.25\\
         
        \rowcolor{light-gray} &Ours &\bf66.37	&\bf174.19	&\bf 38.26	&\bf70.73	&\bf51.61	&\bf53.91	&\bf49.03	&\bf30.72	&\bf36.60 	&\bf47.26

  \\
    \hline
    \end{tabular}
   }
   \caption{Zero-shot Performance of the compressed  LLaMA-2-7B with a pruning rate=50\%. } 
    \label{sec:tbl:v2_result} 
    \vspace{1mm}
\end{table*}
\begin{table*}[h!]
    \centering
    
    \resizebox{0.8\linewidth}{!}{ 
    \small
    \begin{tabular}{c|cc|ccccccc|c}
        \hline
       \bf Ratio &\bf WikiText2$\color{teal}\downarrow$ & \bf PTB$\color{teal}\downarrow$ &\bf BoolQ & \bf PIQA & \bf HellaSwag &\bf WinoGrande &\bf ARC-e & \bf ARC-c &\bf OBQA & \textbf{Average$\uparrow$} \\
       \hline
        \multirow{2}{*}{5\%} &16.91	&58.73 	&77.98	&77.58	&71.01	&67.25	&53.83	&39.85	&41.60	&61.03  \\
           &16.75	&58.80	&72.97	&77.80	&71.56	&68.75	&57.74	&41.21	&42.60 	&61.80   \\
        \hline
        \multirow{2}{*}{10\%} &18.25	&63.83	&75.99	&76.12	&69.96	&67.40 	&54.00 	&39.85	&40.40	&60.53   \\
              &17.69	&61.38	&74.77	&76.71	&76.71	&68.27	&54.50	&38.57	&42.40	&61.70   \\
        \hline 
        \multirow{2}{*}{15\%} &19.17	&66.12	&71.31	&76.99	&70.10	&67.48	&55.18	&40.02	&40.4	&60.21  \\
             &19.35	&65.86	&72.32	&76.66	&70.00	&67.56	&56.65	&39.76	&41.8	&60.68  \\
        \hline 
        \multirow{2}{*}{20\%} &21.68	&72.89	&70.46	&76.22	&68.47	&66.14	&53.24 	&38.23	&41.60	&59.19   \\
            & 21.63	&72.61	&70.85	&77.19	&70.12	&67.16	&53.97	&40.13	&42.06	&60.21  \\

    \hline
    \end{tabular}
   }
   \caption{Pruning ratio for Vicuna-7B with number of samples=50.} \label{sec:tbl:vicuna_ratio}
   \vspace{1mm}
\end{table*}

\subsection{H.2 \quad Compare Adaptive Fusion Estimation with Fix Fusion Estimation}
In Table~\ref{sec:tbl:adap_abl}, Table~\ref{sec:tbl:adap_abl2} we maintained a fixed fusion rate of 0.5 throughout the experiment. For comparison with the adaptive fusion method, we multiplied the coarse-grained evaluation and the fine-grained evaluation by the fusion rate separately and then combined them. The results revealed that the average accuracy of the adaptive fusion method was around 1.4\% higher than that of the fixed fusion method. This highlights the superiority of the adaptive fusion approach in achieving better performance.\\

\subsection{H.3 \quad Compare grouping with no grouping}
In Table~\ref{sec:tbl:llama7B_tune_group}, we present the performance of the LLaMA-7B model across multiple tasks under different pruning ratios and grouping conditions. The table includes three scenarios: no pruning (Ratio = 0\%), 20\% pruning without fine-tuning (Ratio = 20\% w/o tune), and 20\% pruning with LoRA fine-tuning (Ratio = 20\% w/ LoRA). The grouped pruning method (Grouped) significantly outperforms the non-grouped pruning method (No Group) across multiple tasks. Especially when combined with LoRA fine-tuning, the grouped pruning method's effectiveness is comparable to or exceeds the performance of the unpruned model.

\section{I \quad More Ablation Study for Vicuna-7B}
We present the results of the ablation study conducted on Vicuna-7B in Table~\ref{sec:tbl:vicuna_ratio} and Table~\ref{sec:tbl:vicuna_num}.

\begin{table*}[h!]
    \centering
    
    \resizebox{0.8\linewidth}{!}{ 
    \small
    \begin{tabular}{c|cc|ccccccc|c}
        \hline
        \bf Length &\bf WikiText2$\color{teal}\downarrow$ & \bf PTB$\color{teal}\downarrow$ &\bf BoolQ &\bf PIQA &\bf  HellaSwag &\bf WinoGrande &\bf ARC-e & \bf ARC-c &\bf OBQA & \bf Average Accuracy\\
         \hline
        \multirow{2}{*}{10} &21.55	&73.47  &53.79	&77.37	&68.44	&64.72	&54.17	&38.48	&41.20	&56.88  \\
           &21.85	&74.33  &60.53	&76.12	&68.42	&63.93	&57.06	&38.74	&40.60	&57.91   \\
        \hline
       \multirow{2}{*}{20}  &21.85	&75.80 &72.20	&76.17	&68.92	&65.19	&54.63	&38.57	&40.80	&59.49  \\
            &21.88	&74.63	&72.78	&76.33	&68.74	&64.56	&64.29	&38.14	&42.00 	&60.98  \\
        \hline
        \multirow{2}{*}{30} &22.15	&74.33	&69.42	&76.50	&68.42	&66.77	&54.42 	&37.97	&40.80	&59.19 \\
           &22.05	&73.46	&71.74	&76.82	&68.33	&65.51	&54.00 	&37.80	&42.00 	&59.45  \\
       \hline
        \multirow{2}{*}{40} & 22.41	&74.05	&71.87	&76.61	&68.62	&65.51	&54.88 	&39.08	&40.40	&59.56 \\
           &22.62	&74.63	&72.20	&76.28	&68.58	&66.38	&53.32	&38.65	&41.00 	&59.49  \\
         \hline
        \multirow{2}{*}{50} &21.68	&72.89	&70.46	&76.22	&68.47	&66.14	&53.24 	&38.23	&41.60	&59.19 \\
           &\bf 21.63	&\bf 72.61	&\bf 70.85	&\bf 77.19	&\bf 70.12	&\bf 67.16	&\bf 53.97	&\bf 40.13	&\bf 42.06	&\bf 60.21  \\
     \hline
    \end{tabular}
    }
   \caption{Sample numbers for Vicuna-7B with a pruning rate=20\%.} \label{sec:tbl:vicuna_num}
\end{table*}

\begin{table*}[h]
    \vspace{0mm}
    \centering
    
    \resizebox{\linewidth}{!}{
    \begin{tabular}{c|l|cc|ccccccc|c}
        \hline
        \textbf{Ratio} &\textbf{Metho}d & \textbf{WikiText2$\downarrow$} &  \textbf{PTB$\downarrow$} &  \textbf{BoolQ} &  \textbf{PIQA} &  \textbf{HellaSwag} &  \textbf{WinoGrande} &  \textbf{ARC-e} &  \textbf{ARC-c} &  \textbf{OBQA} &  \textbf{Average$\uparrow$}   \\
        \hline
        
        \multirow{2}{*}{20\%}
        &No Adaptive &17.12	&31.83	&67.16	&78.07	&70.11	&64.17	&68.22	&39.51 	&41.20 	&61.20\\

        &Adaptive  &\bf16.42	&\bf31.16 &\bf68.53	&\bf77.80	&7\bf0.58	&\bf67.49	&\bf70.24	&\bf40.44 	&\bf42.00 	&\bf62.44 \\ 
    \hline
    \end{tabular}
   }
   \caption{Adaptive estimation for LLaMA-7B with a pruning rate=20\%. } 
    \label{sec:tbl:adap_abl} 
  
\end{table*}

\begin{table*}[h]
    \vspace{0mm}
    \centering
    
    \resizebox{\linewidth}{!}{
    \begin{tabular}{c|l|cc|ccccccc|c}
        \hline
        \textbf{Ratio} &\textbf{Metho}d & \textbf{WikiText2$\downarrow$} &  \textbf{PTB$\downarrow$} &  \textbf{BoolQ} &  \textbf{PIQA} &  \textbf{HellaSwag} &  \textbf{WinoGrande} &  \textbf{ARC-e} &  \textbf{ARC-c} &  \textbf{OBQA} &  \textbf{Average$\uparrow$}   \\
        \hline
        
        \multirow{2}{*}{50\%}
        &No Adaptive &30.13	&45.10	&60.42	&69.48	&53.76	&53.04	&50.93	&29.27 	&36.10 	&50.42\\

        &Adaptive  &\bf29.35	&\bf44.38 &\bf60.55	&\bf72.36	&\bf 55.25	&\bf55.09	&\bf50.84	&\bf31.48 	&\bf37.00 &\bf51.80 \\
    \hline
    \end{tabular}
   }
   \caption{Adaptive estimation for LLaMA-7B with a pruning rate=50\%. } 
    \label{sec:tbl:adap_abl2} 
\end{table*}

\begin{table}[t]
    \vspace{1mm}
    \centering
   
    \resizebox{\linewidth}{!}{
    \begin{tabular}{cl|cc|ccccccc|c}
        \toprule
        \bf Pruning Ratio &\bf Method &\bf WikiText2$\color{teal}\downarrow$ & \bf PTB$\color{teal}\downarrow$ &\bf BoolQ &\bf PIQA &\bf HellaSwag &\bf WinoGrande &\bf ARC-e &\bf ARC-c &\bf OBQA &\bf Average \\
        \midrule
         Ratio = 0\% & LLaMA-7B& 12.62 & 22.14 & 73.18 & 78.35 & 72.99 & 67.01 & 67.45 & 41.38 & 42.40 & 63.25 \\
        \cmidrule{1-12}
        \multirow{3}{*}{Ratio = 20\% w/o tune} & No Grouped & 61.12 & 73.20& 61.62	&70.40	&48.28	&56.12	&50.42	&30.12	&34.40	&50.19 \\
       
        &Grouped & \bf18.27 &\bf35.16 &\bf65.84	&\bf76.77	&\bf67.87	&\bf60.06	&\bf64.90 	&39.33 	&\bf39.60 	&\bf59.20\\
       
        \cmidrule{1-12}
        \multirow{3}{*}{Ratio = 20\% w/ LoRA} & No Grouped & 21.78 & 38.64 & 61.89 & 70.81 & 58.34 & 56.87 & 54.87 & 34.02 & 38.40 & 53.59 \\
       
        & Grouped  &\bf16.42	&31.16 &\bf68.53	&\bf77.80	&\bf70.58	&\bf67.49	&\bf70.24	&\bf40.44 	&\bf42.00 	&\bf62.44\\ 
     
        \bottomrule
    \end{tabular}
    }
    \vspace{0mm}
    \caption{Compare grouping with no grouping for LLaMA-7B with a pruning rate=20\%. 
    }  \label{sec:tbl:llama7B_tune_group}
    \vspace{3mm}
\end{table}

\section{J \quad Performance Analyze}
We provide ablation experiments with performance analysis with or without fine-tuning in Table~\ref{sec:tbl:llama7B_tune}, Table~\ref{sec:tbl:vica7B_tune}. These results demonstrate the effectiveness of the Adaptive method in pruning large models, providing a superior balance of performance and efficiency compared to traditional coarse and fine-grained pruning techniques.
\begin{table}[t]
    \vspace{1mm}
    \centering
   
    \resizebox{\linewidth}{!}{
    \begin{tabular}{cl|cc|ccccccc|c}
        \toprule
        \bf Pruning Ratio &\bf Method &\bf WikiText2$\color{teal}\downarrow$ & \bf PTB$\color{teal}\downarrow$ &\bf BoolQ &\bf PIQA &\bf HellaSwag &\bf WinoGrande &\bf ARC-e &\bf ARC-c &\bf OBQA &\bf Average \\
        \midrule
         Ratio = 0\% & LLaMA-7B& 12.62 & 22.14 & 73.18 & 78.35 & 72.99 & 67.01 & 67.45 & 41.38 & 42.40 & 63.25 \\
        \cmidrule{1-12}
        \multirow{3}{*}{Ratio = 20\% w/o tune} & Coarse & 22.28 & 41.78 & {61.44} & 71.71 & 57.27 & 54.22 & 55.77 & 33.96 & 38.40 & 53.25 \\
        & Fine  & 24.70 & 94.34 & 62.87 &75.41 &64.00 & 58.41 & 60.98 & 37.12 & 39.00 & 56.83 \\
        & Ours & \bf18.27 &\bf35.16 &\bf65.84	&\bf76.77	&\bf67.87	&\bf60.06	&\bf64.90 	&39.33 	&\bf39.60 	&\bf59.20\\
       
        \cmidrule{1-12}
        \multirow{3}{*}{Ratio = 20\% w/ LoRA} & Coarse  &19.94 &74.66 & 63.15 & 74.59 & 61.95 & 60.30 & 60.48 & 36.60 & 39.40 & 56.64 \\
        & Fine & 18.97 & 76.78 & 60.40 & 75.63 & 65.45 & 63.22 &  63.05 & 37.71 & 39.00 & 57.78 \\
        & Ours &\bf16.42	&31.16 &\bf68.53	&\bf77.80	&\bf70.58	&\bf67.49	&\bf70.24	&\bf40.44 	&\bf42.00 	&\bf62.44\\ 
     
        \bottomrule
    \end{tabular}
    }
    \vspace{0mm}
    \caption{Zero-shot performance of the compressed LLaMA-7B. Here we compared the Corase-grained~\citep{ma2023llm}, Fine-grained~\citep{ma2023llm} , and Our Adaptive method. 
    }  \label{sec:tbl:llama7B_tune}
\end{table}

\begin{table}[t]
    \vspace{1mm}
    \centering
   
    \resizebox{\linewidth}{!}{
    \begin{tabular}{cl|cc|ccccccc|c}
        \toprule
        \bf Pruning Ratio &\bf Method &\bf WikiText2$\color{teal}\downarrow$ & \bf PTB$\color{teal}\downarrow$ &\bf BoolQ &\bf PIQA &\bf HellaSwag &\bf WinoGrande &\bf ARC-e &\bf ARC-c &\bf OBQA &\bf Average \\
        \midrule
         Ratio = 0\% & Vicuna-7B & 16.11 & 61.37 & 76.57 & 77.75 & 70.64 & 67.40 & 65.11 & 41.21 & 40.80 & 62.78 \\
        \cmidrule{1-12}
        \multirow{3}{*}{Ratio = 20\% w/o tune} & Coarse &27.03 & 92.51 & 62.17 & 71.44 & 55.80 & 53.43 & 55.77 & 33.28 & 37.80 & 52.81 \\
        & Fine  & 19.77 & 36.66 & 59.39 & 75.57 & 65.34 & 61.33 & 59.18 & 37.12 & 39.80 & {56.82} \\
        & Ours & \bf17.36 &\bf35.23 &\bf65.81	&\bf76.57	&\bf67.93	&\bf61.32	&\bf65.10 	&39.56 	&\bf39.20 	&\bf59.35\\
       
        \cmidrule{1-12}
        \multirow{3}{*}{Ratio = 20\% w/ LoRA} & Coarse & 18.84 & 33.05 & 65.75 & 74.70 & 64.52 & 59.35 & 60.65 & 36.26 & 39.40 & 57.23\\
        & Fine & 17.37 & 30.39 & 69.54 & 76.44 & 68.11 &65.11 & 63.43 & 37.88 & 40.00 & 60.07 \\
        & Ours &\bf 16.63	&\bf 29.61	&\bf72.78	&\bf 77.33	&\bf68.74	&\bf66.56	&\bf64.29	&\bf38.14	&\bf42.00	&\bf61.35\\
     
        \bottomrule
    \end{tabular}
    }
    \vspace{-3mm}
    \caption{Zero-shot performance of the compressed Vicuna-7B. Here we compared the Corase-grained~\citep{ma2023llm}, Fine-grained~\citep{ma2023llm} , and Our Adaptive methods. 
    }  \label{sec:tbl:vica7B_tune}
\end{table}

\begin{table*}[h!]
    \vspace{-0mm}
    \centering
    
    \resizebox{\linewidth}{!}{
    \begin{tabular}{c|ccccccccccccccccccccccccc}
       \hline
       Method & 5& 6& 7 &8 &9 &10 &11 &12 &13 &14 &15 &16 &17 &18 &19 &20 &21 &22 &23 &24 &25 &26 &27 &28 &29        \\
        \hline
        Original      &135.27 &135.27  &135.27 &135.27 &135.27 &135.27  &135.27 &135.27  &135.27 &135.27  &135.27 &135.27 &135.27 &135.27  &135.27 &135.27 &135.27 &135.27  &135.27 &135.27 &135.27 &135.27  &135.27 &135.27 &135.27\\
        \hline
        Fine &134.75 &133.85 &131.57 &126.81 &118.12 &105.17 &88.92 &80.77 &70.96 &62.16 &58.42 &53.8 &50.27 &46.74 &42.59 &39.46 
        &35.4 &32.64 &29.15 &26.7 &24.64 &24.74 &27.02 &32.34 &41.16\\
        Coares &55.23 &54.86 &68.9 &66.51 &55.0 &50.17 &57.4 &54.53 &67.28 &65.04 &67.74 &60.89 &64.98 &60.48 &48.94 &45.24 &40.66 &54.37 &59.55 &65.9 &75.01 &94.98 &104.63 &110.22 &120.64\\
        Adaptive &90.17 &85.49 &82.93 &79.59 &78.72 &74.15 &73.22 &71.14 &69.19 &69.01 &68.77 &61.25 &59.83 &60.26 &54.19 &53.44 &59.4 &46.03 &53.28 &52.24 &50.18 &50.95 &52.51 &61.01 &75.12\\

       
    \hline
    \end{tabular}
    }
   \caption{The LLaMA-7B adaptive pruning method, with a pruning rate of 50\%, was compared against the fine-grained estimation and coarse-grained estimation pruning methods for parameter quantities in layers 5-29.} \label{sec:compare_num}
\end{table*}

\section{K \quad Sensitivity Analysis for Adaptive Fusion Network. }
We analyzed the computational overhead through Algorithm ~\ref{alg:resource_measurement}: time spent, amount of memory consumed. We measured the memory usage of the Adaptive Fusion network on NVIDIA GPUs to be between 1.04 MB and 3.00MB, it takes about 0.013970 seconds.
\begin{algorithm}[h]
\caption{Resource Usage Measurement for Adaptive Fusion}
\label{alg:resource_measurement}
\textbf{Input}: Fine-grained gradients, Coarse-grained gradients\\
\textbf{Output}: Memory usage \texttt{mem\_use}, Time usage \texttt{time\_use}
\begin{algorithmic}[1] 
\STATE \texttt{start\_time} $\leftarrow$ \texttt{time.time()}
\STATE \texttt{start\_mem} $\leftarrow$ \texttt{memory\_usage()}

\STATE \texttt{adaptive\_fuse(fine\_grained\_grad, coarse\_grained\_grad)}

\STATE \texttt{end\_time} $\leftarrow$ \texttt{time.time()}
\STATE \texttt{end\_mem} $\leftarrow$ \texttt{memory\_usage()}

\STATE \texttt{mem\_use} $\leftarrow$ \texttt{end\_mem - start\_mem}
\STATE \texttt{time\_use} $\leftarrow$ \texttt{end\_time - start\_time}

\STATE \textbf{return} \texttt{mem\_use}, \texttt{time\_use}
\end{algorithmic}
\end{algorithm}

\section{L \quad Generations From Compressed Model}
Table~\ref{generate_com} resent examples of the models pruned by our method. We show the generation results of dense models and various pruning methods.
\begin{table*}[h!]
\centering
\vspace{-0.05in}
\begin{adjustbox}{width=\columnwidth}
\begin{threeparttable}
\begin{tabular}{p{1.8cm}|p{14.5cm}}

\specialrule{.2em}{.1em}{.1em} 
Model       & Generated Text 
\\ \hline

\small{LLaMA-7B}    & 
\footnotesize \underline{What’s great about the holiday season,} it’s a time when people focus less on work and more on families and friends. The holidays are a time when people can really focus on giving and helping others, which is something that the holidays are supposed to be about. In my blog post this week, I’ll be taking a look at a variety of causes that ...
\\ \hline

\small{Wanda-sp}    & 
\footnotesize \underline{What’s great about the holiday season,} though, is its warm and inviting people, and what a chance to meet and share a cuppa with others who understand and have shared similar experiences, a feeling of empathy if you will, and a great opportunity to help others on a difficult journey. So I can say, all is well. But I'm not giving up the fight ...
\\ \hline
 \small{FLAP}        & 
\footnotesize \underline{What’s great about the holiday season,} and what makes it such an awesome time of the year, is that it’s the time of the year when people are so generous, and so goodhearted. When you get together with friends and family, and you go through the various things that you’re able to do, when you’re able to give people gifts, when people can see ...
\\ \hline
 \small{LLM-Pruner}  & 
\footnotesize \underline{What’s great about the holiday season,} is you can enjoy the company of friends and family. This is the perfect opportunity to gather together, and share all the special memories that you have together. These moments are the ones that make the holiday special. No matter what happens, or what you end up doing, it’s always good to get ...
\\ \hline
\small{Ours}  & 
\footnotesize \underline{What’s great about the holiday season,} is the way it brings so many people together, no matter how close or far apart they live. So don’t hesitate to reach out and let someone know how much you appreciate them this year. Whether it’s a phone call, a message, a visit, or a simple handshake, your gratitude for that person’s existence in your life is all they need to hear...
\\ 
\specialrule{.2em}{.1em}{.1em}

\small{LLaMA-7B}    & 
\footnotesize \underline{AI can create a logo in seconds}. Why are these new technologies threatening our jobs? AI can do most menial tasks in just a few seconds or minutes. These tasks can be very tedious for human beings to do. For instance, AI can identify patterns in large amounts of data so that complex statistical analysis can be carried out faster than a ...
\\ \hline

\small{Wanda-sp}    & 
\footnotesize \underline{AI can create a logo in seconds} rather than years. To this point, this is only the beginning of what AI will do to change our human lives. As time goes on, it’s quite likely that we’ll see even greater use of robots. Even if this all sounds too dramatic, this may still happen. And AI is the best thing that can make it to happen. in 2019. All you ...
\\ \hline

 \small{FLAP}        & 
\footnotesize \underline{AI can create a logo in seconds} and it doesn’t cost you a penny. The new era of technology has opened up many exciting opportunities. Technology, like AI is being utilized for different purposes in different ways, and now, technology is becoming a solution to problems that would have been tedious or expensive only a few years ago. ...
\\ \hline
 \small{LLM-Pruner}  & 
\footnotesize \underline{AI can create a logo in seconds}, design a website from scratch and even write blog posts without any human intervention. This revolution is changing the way humans communicate and express ideas. In our view, this technology is a great asset but we'll need to be smart in how we use it to prevent abuse or automation leading to ...
\\ \hline

 \small{Ours} & 
\footnotesize \underline{AI can create a logo in seconds}, revolutionizing the design process by offering speed, convenience, and cost-effectiveness. By leveraging machine learning algorithms, AI-powered design tools analyze millions of design elements from fonts and colors to shapes and styles to generate a variety of logo options tailored to specific preferences or industries...
\\ 
\specialrule{.2em}{.1em}{.1em}

\small{LLaMA-7B}    & 
\footnotesize \underline{Neural network pruning is defined as} the process of removing certain weights or neurons in a neural network to reduce the size of the neural network, and improve the training process. Neural networks pruning techniques have been applied successfully on classification, optimization, planning, and speech recognition tasks. A pruned ...
\\ \hline
\small{Wanda-sp}    & 
\footnotesize \underline{Neural network pruning is defined as} any method that removes the redundant information by applying the neural network network. It is the term used to identify the systematic variables when an individual tries to find an algorithm. In addition, it is the method of classification used to define any set of algorithms that have been ...
\\ \hline
 \small{FLAP}        & 
\footnotesize \underline{Neural network pruning is defined as} a selection of neurons or weights in a neural network so that the information of the network is reduced to a specified level, and a process of re-training a neural network. The latter one is the most popular method, especially the neural network pruning. In recent years, this field has been studied in ...
\\ \hline
 \small{LLM-Pruner}  & 
\footnotesize \underline{Neural network pruning is defined as} reducing the complexity and size of an over-parameterized neural network. The pruning reduces the FLOPs and improves performance for the particular task. Pruning techniques have gained popularity in recent years due to their wide range of applications. This paper aims to explore the pruning ...
\\ \hline
\small{Ours}        & 
\footnotesize \underline{Neural network pruning is defined as} the process of selecting a subset of connections within a neural network to keep the network performant, while maintaining or even improving the accuracy. In this context, optimization algorithms are used to reduce the weights of the unimportant connections, thereby achieving higher accuracy with fewer resources, particularly in the case of Deep Learning techniques...\\
\specialrule{.2em}{.1em}{.1em}
\end{tabular}
\end{threeparttable}
\end{adjustbox}
\vspace{-0.05in}
\caption{Generation examples from the original LLaMA-7B and 20\%-compressed models.} \label{generate_com}
\end{table*}


\end{document}